\documentclass[a4paper]{article}[12pts]

\usepackage[english]{babel}
\usepackage[utf8x]{inputenc}
\usepackage[T1]{fontenc}

\normalsize

\usepackage[a4paper,top=1in,bottom=1in,left=1in,right=1in,marginparwidth=1in]{geometry}

\usepackage{setspace}
\usepackage{amsmath}
\usepackage{amssymb} 
\usepackage{amsthm}
\usepackage{amsfonts, mathtools}
\usepackage{algorithm,algpseudocode}
\usepackage{bm}
\usepackage{bbm}
\usepackage{comment}
\usepackage{graphicx}
\usepackage{multirow, booktabs}
\usepackage{caption}
\usepackage{subcaption}
\usepackage[colorinlistoftodos]{todonotes}
\usepackage[bottom]{footmisc}
\usepackage[colorlinks=true, allcolors=blue,hypertexnames=false]{hyperref}
\usepackage{pifont}

\DeclareMathOperator*{\argmin}{arg\,min}

\theoremstyle{plain}
\newtheorem{proposition}{Proposition}

\newtheorem{assumption}{Assumption}
\newtheorem{theorem}{Theorem}
\newtheorem{corollary}{Corollary}

\theoremstyle{definition}
\newtheorem{definition}{Definition}
\newtheorem{example}{Example}

\theoremstyle{remark}
\newtheorem{remark}{Remark}

\newcommand{\E}{\mathbb{E}}

\usepackage{natbib}
\usepackage{adjustbox}

\usepackage{placeins}
\allowdisplaybreaks
\hypersetup{pdftitle={Understanding Uncertainty Sampling via Equivalent Loss},
  pdfauthor={Shang Liu and Xiaocheng Li}}

\title{Understanding Uncertainty Sampling via Equivalent Loss}
\author{Shang Liu\textsuperscript{1,}\thanks{Corresponding author: Shang Liu. Email: \href{mailto:s.liu21@imperial.ac.uk}{\texttt{s.liu21@imperial.ac.uk}}.}\quad Xiaocheng Li\textsuperscript{2}}
\date{\small
\textsuperscript{1}Imperial College Business School, Imperial College London\\
\textsuperscript{2}Uber}

\begin{document}
\maketitle
\onehalfspacing

\begin{abstract}
Uncertainty sampling is a classical active-learning strategy, yet the statistical objective induced by its query rule is often implicit. We introduce the equivalent loss, whose gradient is the original loss gradient multiplied by the query probability. This construction places probabilistic, margin-based, and threshold-based uncertainty rules within a common framework. For binary classification, concrete equivalent losses and surrogate link functions show how uncertainty weighting preserves calibration while reshaping optimization geometry. When the equivalent loss is convex, we derive a finite-sample excess-risk bound with a fixed learning rate and an explicit constant controlling the tradeoff between initial error and query-weighted gradient variance. A feasible choice based only on maximal uncertainty yields a leading classification-risk upper bound no larger than the corresponding passive-learning bound at the same expected label budget; knowledge of the average query rate sharpens this comparison. We also analyze pool-based sampling, characterize the integrability obstruction beyond scalar prediction, and examine the query-clock dynamics of momentum methods. Together, these results provide a reusable route from a classical acquisition rule to its induced objective, statistical guarantees, and optimization behavior.
\end{abstract}


\section{Introduction}
\label{sec:intro}

Active learning allows a learner to decide which observations should be labeled. Unlike ordinary supervised learning, it begins with unlabeled features and acquires labels adaptively. This setting is especially attractive when unlabeled data are abundant but annotation is expensive. The objective is to concentrate a limited labeling budget on observations that are informative for the task.

Uncertainty sampling is one of the oldest and most widely used acquisition rules: it favors observations on which the current model is least confident \citep{lewis1995sequential,settles2009active}. Its appeal lies in connecting prediction confidence directly to the acquisition decision. The same principle appears in entropy and least-confidence rules for probabilistic models, distance-to-boundary rules for margin classifiers, and hard thresholds that query only near a decision boundary. These different forms are easy to implement, but they do not obviously share a common statistical objective.

Existing analyses illuminate important special cases. Threshold-based sampling can be interpreted as a preconditioned stochastic-gradient method on a smoothed $0$--$1$ objective, exposing the role of non-convexity and initialization \citep{mussmann2018uncertainty}. Margin-based procedures admit convergence results under strict linear separability \citep{raj2022convergence}. Conversely, high-dimensional constructions demonstrate that least-confidence sampling can perform worse than passive sampling \citep{tifrea2022uniform}. Query-synthesis analyses establish favorable constants under specific distributional assumptions \citep{mussmann2022constants}. Despite this progress, a general route from a classical uncertainty score to a label-dependent performance analysis remains scarce.

The starting point of this paper is that randomized querying changes the expected gradient field. If a feature $X$ is queried with probability $U(\theta;X)$ and an accepted label produces the gradient of $\ell(\theta;(X,Y))$, then averaging over the query decision produces $U\nabla_\theta\ell$. Whenever this field is conservative, it is the gradient of an \emph{equivalent loss} $\tilde\ell$. The acquisition rule can therefore be studied through an explicit transformed objective. This distinguishes three questions: which objective the queried direction represents, whether that objective remains calibrated for the task, and whether its geometry supports efficient optimization.

To our knowledge, the equivalent-loss construction provides the first unified objective-based framework for analyzing these classical uncertainty-sampling rules and relating their statistical behavior to label complexity. The framework is useful precisely because it applies across rules rather than prescribing one score. The entropy, least-confidence, smooth-margin, and threshold examples below show how previously separate algorithms enter the same calculation; the resulting losses reveal both positive cases and genuine obstructions.

Our contributions have four components. First, we identify the equivalent loss and derive it explicitly for representative uncertainty--loss pairs. Second, we use surrogate-risk theory to study calibration and risk transfer, while a separate curvature calculation explains when uncertainty weighting creates non-convexity. Third, we derive a label-dependent convergence bound for a fixed step size $\eta_t=cD/(G\sqrt T)$ and translate it into excess binary risk through the surrogate link. The constant $c$ exposes the tradeoff between initial error and the variance of queried gradients. Choosing $c=U_{\text{max}}^{-1/2}$ uses only the known maximum query probability and gives a leading upper bound no larger than the passive-learning counterpart at the same expected label budget. An oracle balance using the average query rate clarifies the additional benefit of more accurate rate calibration. Fourth, we extend the viewpoint to pool-based sampling and examine two boundaries: integrability beyond scalar prediction and the optimizer clock for momentum methods.

The broader contribution is a reusable analytical route. Identifying the induced vector field reveals which risk is relevant before one applies an optimization theorem. Studying the corresponding loss then exposes whether the algorithm retains a statistically meaningful target and whether selective querying has altered its landscape. These quantities give readers a way to compare acquisition rules, choose a learning-rate scale, and connect optimization guarantees to label complexity.

\subsection{Related Works}
\paragraph{Surrogate losses and calibration.}
Surrogate losses replace task losses such as the $0$--$1$ loss with objectives that are computationally tractable. Classification calibration, also called Fisher consistency, ensures that population minimization recovers the Bayes decision \citep{zhang2004statistical,bartlett2006convexity,scott2011surrogate}. The optimal link function of \citet{bartlett2006convexity} strengthens this consistency statement by converting surrogate excess risk into excess binary risk. Proper composite losses characterize probability estimation \citep{reid2010composite}, while multiclass calibration and convex calibration dimension describe the geometric requirements beyond binary prediction \citep{tewari2007consistency,ramaswamy2016convex}. Our analysis applies these tools after incorporating the acquisition rule into the equivalent loss.

\paragraph{Active-learning theory.}
Disagreement, version-space, and compression methods provide label-complexity guarantees under different structural assumptions \citep{balcan2006agnostic,hanneke2014theory,wiener2015compression,hanneke2015minimax}. Results for uncertainty sampling itself are more specialized. \citet{mussmann2018uncertainty} connect threshold sampling to a smoothed classification objective; \citet{raj2022convergence} analyze a stream-based margin rule under separability; and \citet{tifrea2022uniform} exhibit an adverse high-dimensional instance. \citet{mussmann2022constants} study logistic regression in a query-synthesis model. Following the initial version of this work, \citet{haimovich2024convergence} build on the equivalent-loss construction to obtain convergence guarantees for a broader family of loss-based sampling rules and develop adaptive-weight sampling. These developments connect acquisition design with the objective seen by optimization. Our analysis brings that objective together with calibration, geometry, and label complexity; Appendix~\ref{apd:additional-related-work} discusses the broader context.

\section{Problem Setup}
\label{sec:setup}
Consider the problem of predicting the label $Y$ from the feature $X$, where $(X, Y)$ is independently drawn from an unknown distribution $\mathcal{P}$. We denote the marginal distribution of $X$ to be $\mathcal{P}_X$ and the conditional distribution of $Y$ on $X$ is $\mathcal{P}_{Y|X}$. Let $\mathcal{X}$ and $\mathcal{Y}$ denote the support of $X$ and $Y$ respectively. 
Suppose $\mathcal{X} \subset \mathbb{R}^d$. In this paper, we are mainly on the binary classification problem, where $\mathcal{Y} = \{-1, +1\}$. Later in Section~\ref{sec:multi_dim}, we also consider the $K$-nary classification problem where $\mathcal{Y} = [K] = \{1, \dots, K\}$ and the regression problem where $\mathcal{Y} = \mathbb{R}$. But for now, let's focus on the binary classification problem.

For the canonical setting of supervised learning, a full dataset of both features and labels is completely revealed to the learner at the beginning. For active learning, the learner starts with only observations of the features $X$'s and needs to decide which of the labels $Y$'s to query or whether to query the labels $Y$'s. In this paper, we consider two mainstream settings for active learning.
\begin{itemize}
\item \textbf{Stream-based setting.} The dataset $\mathcal{D}_T^X$ consists of $T$ i.i.d. features $\{X_t\}_{t=1}^T$ from $\mathcal{P}_X$. The samples arrive sequentially. At each time $t$, upon the arrival of $X_t$, the learner decides whether to query the sample: if so, $Y_t$ is revealed to the learner; otherwise, it moves on to the next time period. The feature and the label (if queried) of the $t$-th time period will be discarded (but not cached) after the time period. Without loss of generality, we still assume the presence of the label $Y_t$ sampled from $\mathcal{P}_{Y|X=X_t}$; it may just not be revealed to the learner depending on the querying decision. 
\item \textbf{Pool-based setting.} The dataset $\mathcal{D}_n^{X}$ consists of $n$ i.i.d. features $\{X_i\}_{i=1}^n$ from $\mathcal{P}_X$. The whole dataset $\mathcal{D}_n^{X}$ is revealed all at once to the learner at the beginning. The learner queries samples from the dataset sequentially. Unlike the stream-based setting, the information from past queries will be retained and can be repeatedly utilized by the learner.
\end{itemize}

Throughout the paper, we consider a parameterized family of hypotheses denoted by $\mathcal{F} = \{f_\theta(\cdot): \theta \in \Theta, f_\theta(\cdot):\mathcal{X}\rightarrow \mathbb{R}\}$. Here, we relax the image of $f_\theta$ to $\mathbb{R}$ rather than only $\mathcal{Y}$, since there are many loss functions such as the hinge loss that takes not only the sign of $f_\theta(X)$ but also the absolute value of it as input.
We assume the parameter set $\Theta \subset \mathbb{R}^k$.
We denote the loss function $\ell:\mathbb{R}\times \mathcal{Y} \rightarrow \mathbb{R}$, i.e., $\ell(\hat{Y}, Y)$ measures the loss of predicting $Y$ with $\hat{Y}$. With a slight overload of the notation, we denote $\ell(\theta; (X, Y)) = \ell(f_\theta(X), Y)$ as the prediction loss of the model $f_{\theta}(\cdot)$ on the sample $(X, Y)$.

For uncertainty sampling algorithms, a key component is an uncertainty function/measure $U(\theta; X):\mathcal{X} \rightarrow (0, 1]$. The uncertainty function quantifies the uncertainty about a sample $X$ given the model parameter $\theta$. In this paper, we assume that the uncertainty is almost surely bounded. In the context of uncertainty sampling, this uncertainty also serves as the probability that the label of some sample is queried. Here we assume $U$ to be strictly positive to ease our later analysis. The specification of the uncertainty function usually depends on both the underlying hypothesis class $\mathcal{F}$ and the loss function $\ell$
\citep{dagan1995committee, culotta2005reducing, dasgupta2005analysis, balcan2007margin, mussmann2018uncertainty, raj2022convergence, tifrea2022uniform}. The general idea is to spend more querying efforts on those samples that the current model is uncertain about, in the hope of maximizing the improvement of the model learning. 

\paragraph{Surrogate property.}
Before we dive into the detailed discussions of the uncertainty sampling algorithm, we present the definition of surrogate loss and related results in \citet{bartlett2006convexity}. The introduction of such a surrogate property is to compare which loss functions are ``suitable'' for the binary classification problem. The practical goal of training a binary classifier is commonly to achieve a high classification accuracy, i.e., to optimize the binary loss $\ell_{01}(\hat{Y}, Y) \coloneqq \mathbbm{1}(\mathrm{sgn}(\hat{Y})\neq Y)$. While the binary loss is computationally intractable in general \citep{arora1997hardness}, the margin loss, the logistic loss, and the cross-entropy loss can all be viewed as a \textit{surrogate} loss of the binary loss that enjoys better computational structure, such as convexity.

\begin{definition}[Surrogate loss \citep{bartlett2006convexity}]
A loss function $\ell(\cdot, \cdot)$ is said to be a surrogate of the binary loss if there exists a continuous, non-negative, and non-decreasing function $\psi$ such that for any measurable function $f:\mathcal{X}\rightarrow \mathbb{R}$ and any probability distribution $\mathcal{P}$ on $\mathcal{X} \times \mathcal{Y} = \mathcal{X} \times \{-1, +1\}$,
\begin{equation}
\label{eq:surrogate}
\psi\left(L_{\mathrm{01}}(f) - \inf_{g \in \mathcal{G}} L_{\mathrm{01}}(g) \right) \leq \mathbb{E}\left[\ell(f(X), Y) \right] -\inf_{g \in \mathcal{G}} \mathbb{E}\left[\ell(g(X), Y)\right],
\end{equation}
where $\mathcal{G}$ is the set of all measurable functions, and $L_{\mathrm{01}}(f) \coloneqq \mathbb{E}\left[\ell_{01}(f(X), Y) \right]$ denotes the expected binary loss. All the expectations are taken with respect to the distribution $\mathcal{P}.$ 
\end{definition}

The definition establishes a connection between the oracle generation bound under the loss $\ell$ and that under the binary loss. It can thus verify the properness of a loss $\ell$ by whether training a model with $\ell$ can also lead to a performance guarantee for the binary loss. An important property of the link function $\psi$ is that if $z\rightarrow 0$ as $\psi(z) \rightarrow 0$, then the loss function is classification-calibrated \citep{bartlett2006convexity}. This ensures that the minimizer of the loss $\ell$ among all the measurable functions will be the Bayes optimal classifier; the property is also known as the Fisher consistency. To get a better guarantee on the excess binary loss, we hope to find the optimal link function $\psi$ to be as ``large'' as possible. The following theorem shows how to identify such a minimax optimal $\psi$.

The construction of \citet{bartlett2006convexity} gives a minimax-sharp link in \eqref{eq:surrogate}. Its positivity away from zero characterizes calibration, and its local behavior determines how an optimization rate transfers to classification risk. Appendix~\ref{apd:bartlett} records the precise statement used here.

\citet{bartlett2006convexity} provide a way to derive the link function $\psi$ (See our Appendix \ref{apd:surrogate} for more details). They further prove that this surrogate property's link function is minimax optimal by the existence of a probability distribution to make the surrogate upper bound arbitrarily tight. They also establish some equivalence between the link function and the Fisher consistency. While such a conclusion is only stated for margin-based models where $\ell(\hat{Y}, Y)$ can be expressed as some $\ell(\hat{Y} \cdot Y)$ in \citep{bartlett2006convexity}, their analysis in Theorem \ref{thm:bartlett} indeed applies to more general loss functions such as the cross entropy written as loss $\ell(\hat{Y}, Y) = \mathbbm{1}\{Y=+1\} \cdot \ell(\hat{Y}, +1) + \mathbbm{1}\{Y=-1\} \cdot \ell(\hat{Y}, -1)$.

\section{Uncertainty Sampling for Binary Classification}
\label{sec:binary}

We begin our discussion with the binary classification problem. We first present a generic stream-based algorithm of uncertainty sampling with gradient descent update rules. We then propose a general framework named \emph{equivalent loss} to analyze the properties. As an illustration, we examine the surrogate properties and the convexity conditions for some existing examples.

\subsection{Generic Algorithm under Stream-Based Setting}

Algorithm~\ref{alg:USGD} observes the current feature before deciding whether to query its label. Let
\begin{align*}
\mathcal F_t
&=\sigma\!\left(\theta_1,
\{X_s,\xi_s,\mathbbm{1}\{\xi_s\le U_s\}Y_s\}_{s<t}\right),\\
\mathcal F_t^X&=\mathcal F_t\vee\sigma(X_t).
\end{align*}
Here $\mathcal F_t$ is the observed history before $X_t$ arrives; the product records a label only when it is queried. The fresh pair $(X_t,Y_t)$ is independent of $\mathcal F_t$, and $\xi_t\sim\operatorname{Unif}[0,1]$ is independent of the data and past randomization. Thus
\[
\mathcal F_t\subseteq\mathcal F_t^X\subseteq\mathcal F_{t+1}.
\]
The parameter $\theta_t$ is $\mathcal F_t$-measurable, whereas $U_t=U(\theta_t;X_t)$ is $\mathcal F_t^X$-measurable. In particular, computing uncertainty requires the feature but not the label. A step size depending on the current $U_t$ is also available before the query, although it need not be measurable before $X_t$ arrives. Our convergence results use the fixed choice
\[
\eta_t=\frac{cD}{G\sqrt T},\qquad c>0,
\]
with $c$ selected before the stream is observed. Both the learning rate and the normalizer of the output average are then deterministic.

Conditional on the current observation, averaging the query randomization gives
\begin{equation}
\label{eq:filtration-query}
\E[\theta_{t+1}-\theta_t\mid\mathcal F_t^X\vee\sigma(Y_t)]
=-\eta_tU_t\nabla_\theta\ell(\theta_t;(X_t,Y_t)).
\end{equation}
Conditioning on the latent $Y_t$ is only an analytical device; the algorithm observes it only upon querying. The distinction between $\mathcal F_t$ and $\mathcal F_t^X$ will also let us account for the second moment of the queried gradient.

\begin{algorithm}[tb]
\caption{Uncertainty sampling with gradient descent update (stream-based version)}
    \label{alg:USGD}
    \begin{algorithmic}[1] 
    \Require Dataset $\mathcal{D}_T=\{(X_t, Y_t)\}_{t=1}^{T}$, step sizes $\{\eta_t\}_{t=1}^T>0$ chosen before the current feature, uncertainty function $U:\Theta\times\mathcal X\to[0,1]$
    \State Initialize $\theta_1$; $\bar{\theta}_0 \leftarrow \theta_1$
    \For{$t=1,...,T$}
    \State Observe $X_t$ and calculate $U(\theta_t; X_t)$
    \State Generate $\xi_t\sim \text{Unif}[0,1]$
    \If{$\xi_t \le U(\theta_t; X_t)$}
    \State Query the label $Y_t$ and update
    \[ \theta_{t+1} \leftarrow \theta_t - \eta_t \cdot \frac{\partial \ell(\theta;(X_t, Y_t))}{\partial \theta}\Bigg\vert_{\theta=\theta_t} \]
    \Else 
    \State Do not query the label $Y_t$ and let
    \[ \theta_{t+1} \leftarrow \theta_t\]
    \EndIf
    \State $\bar{\theta}_{t} \leftarrow (1-\frac{\eta_t}{\sum_{s=1}^t \eta_s}) \bar{\theta}_{t-1} + \frac{\eta_t}{\sum_{s=1}^t \eta_s} \theta_{t}$
    \EndFor
    \Ensure $\bar{\theta}_{T}$
    \end{algorithmic}
\end{algorithm}

The core idea of the algorithm is to query only the samples that the model is uncertain about, and the uncertainty function quantifies such uncertainty. In the following, we review three examples of the uncertainty function used in the literature as special cases of the generic algorithm. 

\begin{example}[Probabilistic model]
\label{eg:probabilistic}
A probabilistic model \citep{dagan1995committee, culotta2005reducing} outputs $q(\theta; X): \mathcal{X} \rightarrow [0,1]$ to estimate the true conditional probability $\mathbb{P}(Y=+1|X)$. The entropy uncertainty \citep{dagan1995committee} considers the entropy of $q(\theta; X)$:
\[U(\theta; X) \coloneqq -\left[q(\theta; X) \log(q(\theta; X)) + (1-q(\theta; X))\log (1-q(\theta; X))\right],\]
where $q = q(X;\theta) \in (0, 1)$. The least confidence uncertainty \citep{culotta2005reducing} considers
\[U(\theta; X) \coloneqq 1 - \max\{q(X;\theta), 1-q(X;\theta)\} = \min\{q(X;\theta), 1-q(X;\theta)\}.\]
These two uncertainties are often accompanied by the following cross-entropy loss that trains the probabilistic model
\[\ell(\theta;(X, Y)) = -\left[\mathbbm{1}\{Y=+1\} \log q(\theta;X) + \mathbbm{1}\{Y=-1\} \log(1-q(\theta;X))\right]\]
where $\mathbbm{1}\{\cdot\}$ is the indicator function. Equivalently, we can also represent the loss function by  
$\ell(\hat{Y}, Y) = -\log\left(\frac{1+\hat{Y}\cdot Y}{2}\right)$
where $\hat{Y} = 2 q(\theta; X) - 1 \in [-1, 1]$ is the predicted expectation.
\end{example}

For a probabilistic model, $q(\theta; X)$ reflects the confidence of the prediction. When $q(\theta; X)$ is close to $1$, the model is confident that $Y=+1$, while $q(\theta; X)$ is close to $0,$ it is confident that $Y=-1.$ For both ends, the uncertainty is small for both the entropy uncertainty and the least confidence uncertainty. When the model is less confident about the prediction and outputs $q(\theta; X)$ close to $\frac{1}{2},$ the uncertainty becomes larger.

\begin{example}[Margin-based model]\label{eg:raj&bach} Another class of classification model is margin-based \citep{raj2022convergence}, such as support vector machines (SVMs). Consider a linear SVM model that predicts $Y$ with the sign of $\theta^\top X$. The margin-based uncertainty function is defined by 
\[U_\mu(\theta; X) \coloneqq \frac{1}{1+\mu |\theta^\top X|}\]
where $\mu>0$ is a hyper-parameter. The associated loss function for learning such margin-based models is squared margin loss
\[\ell(\theta;(X, Y)) = (\max\{0, 1-Y\cdot \theta^\top X\})^2.\]
Equivalently, the loss function can be written in the form of
\[\ell(\hat{Y}, Y) = (\max\{0, 1-Y\cdot \hat{Y}\})^2,\]
where $\hat{Y} = \theta^\top X$.
\end{example}

For linear classifiers, $|\theta^\top x|$ is proportional to the distance from a sample to the classification hyperplane. The margin-based uncertainty captures the intuition that the closer a sample is to the classification hyperplane, the more uncertain the learner is about the sample. 

\begin{example}[Threshold-based uncertainty]\label{eg:tifrea} A threshold-based uncertainty function is to only query the samples of which the uncertainty is above a certain threshold \citep{orabona2011better, mussmann2018uncertainty}. The pool-based active learning that always queries the most uncertain sample also results in a threshold-based uncertainty function \citep{tifrea2022uniform}. To see that, at each time step, the algorithm will query the most uncertain sample in the given dataset with index $i_{t} = \argmin_{i\in \mathcal{U}_t} |\theta^{\top} X_i|$ where the set $\mathcal{U}_t$ contains the indices of unqueried samples at time $t$. Such a procedure can be captured by the following uncertainty function
\[U(\theta; X) \coloneqq \mathbbm{1}\{|\theta^{\top} X| \leq \gamma\}.\]
where $\gamma>0$ is a hyper-parameter that may change over time. \citet{tifrea2022uniform} analyze this uncertainty function and derive some negative theoretical results on its performance. Specifically, they consider the following loss for a logistic regression model
\[\ell(\theta;(X, Y)) = \log(1+\exp(-Y \cdot \theta^\top X)).\]
Equivalently, the loss can be written as
\[\ell(\hat{Y}, Y) = \log(1+\exp(-Y \cdot \hat{Y})),\]
where $\hat{Y} = \theta^\top X$.
\end{example}

As in the margin-based model, the quantity $|\theta^\top x|$ reflects the confidence of the prediction, and thus it is inversely proportional to the uncertainty. The threshold-based uncertainty queries only those samples where the confidence is smaller than the threshold $\gamma$.

\subsection{Equivalent Loss}
\label{subsec:equiv_loss}

The equivalent loss identifies the sample-level objective represented by the queried direction. Averaging the two cases in Algorithm~\ref{alg:USGD} over the query randomization, while keeping the current observation fixed, gives
\[ \E_{\xi_t}[\theta_{t+1}|\theta_t,X_t,Y_t] = \theta_t - \eta_t \cdot U(\theta_t; X_t)\cdot \frac{\partial \ell(\theta;(X_t,Y_t))}{\partial \theta}\Bigg\vert_{\theta=\theta_t}\]
The expectation averages only the query randomization, as in \eqref{eq:filtration-query}. The current feature, latent label, and therefore $\eta_t$ are fixed at this stage. 
 
Suppose (for the moment) there exists a loss function $\tilde{\ell}$ such that 
\begin{equation}
\frac{\partial \tilde{\ell}(\theta;(x,y))}{\partial \theta}=U(\theta; X)\cdot \frac{\partial \ell(\theta;(x,y))}{\partial \theta}
\label{eq:key_ODE}
\end{equation}
holds for all $\theta\in\Theta$ and $(x,y)\in \mathcal{X}\times \mathcal{Y}$ (we will discuss the existence of $\tilde{\ell}$ in the following subsection). Then the parameter update can be written as 
\[ \E_{\xi_t}[\theta_{t+1}|\theta_t,X_t,Y_t] = \theta_t - \eta_t \cdot \frac{\partial \tilde{\ell}(\theta;(X_t,Y_t))}{\partial \theta}\Bigg\vert_{\theta=\theta_t}.\]

\begin{proposition}
Suppose there exists $\tilde{\ell}$ satisfying \eqref{eq:key_ODE}. Then the query-conditional expected update of Algorithm~\ref{alg:USGD} equals a gradient step on $\tilde\ell$:
\[ \E_{\xi_t}[\theta_{t+1}|\theta_t,X_t,Y_t] - \theta_t = - \eta_t \cdot \frac{\partial \tilde{\ell}(\theta;(X_t,Y_t))}{\partial \theta}\Bigg\vert_{\theta=\theta_t}.\]
\label{prop:SGD_equiv}
\end{proposition}

\begin{definition}
We say $\tilde{\ell}$ is the \textit{equivalent loss} for the uncertainty function $U$ and the original loss function $\ell$, if it satisfies \eqref{eq:key_ODE}.
\end{definition}

The equivalent loss turns the acquisition mechanism into an objective that can be studied using surrogate-risk theory and stochastic optimization. Write the queried direction as
\[
g_t=\mathbbm{1}\{\xi_t\le U_t\}\nabla_\theta\ell(\theta_t;(X_t,Y_t)).
\]
Successively averaging the query decision, the label, and the feature gives
\begin{align}
\E[g_t\mid\mathcal F_t^X]
&=U_t\E[\nabla_\theta\ell(\theta_t;(X_t,Y_t))\mid\mathcal F_t^X],\label{eq:feature-gradient}\\
\E[g_t\mid\mathcal F_t]
&=\nabla_\theta\E_{(X,Y)\sim\mathcal P}[\tilde\ell(\theta_t;(X,Y))].\label{eq:history-gradient}
\end{align}
For an $\mathcal F_t$-measurable step size, the conditional mean update is therefore $-\eta_t$ times the population equivalent-risk gradient. The fixed step used below is one such choice. These identities explain how the sample-level representation becomes a population optimization guarantee; the explicit losses for the three running examples follow next.

\setcounter{example}{0}
\begin{example}[Continued]
Example 1 considers a probabilistic model $q(\theta; X)$ that estimates the true conditional probability $\mathbb{P}(Y=+1|X)$, and the loss function is the cross-entropy loss.
\begin{itemize}
\item For the entropy uncertainty, the equivalent loss
\begin{align*}
\tilde{\ell}(\theta; (X,Y)) & = 
q\log(q) + (1-q)\log(1-q) \\
& \phantom{=} - \mathbbm{1}\{Y=+1\} \cdot \mathrm{Li}_2(q) - \mathbbm{1}\{Y=-1\} \cdot \mathrm{Li}_2(1-q) + \mathrm{Li}_2(1),
\end{align*}
where $q=q(\theta;X)$ stands for the prediction model and the function 
\[\mathrm{Li}_2(z) = -\int_0^z \frac{\log(1-u)}{u} \mathrm{d}u\]
is Spence's function.
\item For the least confidence uncertainty, the equivalent loss
\[\tilde{\ell}(\theta; (X,Y)) = \begin{cases}
-\mathbbm{1}\{Y=-1\}\cdot\log(2(1-q)) - q + \log(2), \quad & \text{if } q < \frac12;\\
-\mathbbm{1}\{Y=+1\}\cdot\log(2q) - (1-q) + \log(2), \quad & \text{if } q \geq \frac12.
\end{cases}\]
where $q=q(\theta;X)$ stands for the prediction model.
\end{itemize}

\end{example}

\begin{example}[Continued] For the margin-based model, the equivalent loss for the margin-based uncertainty function (defined in Example 2) and the squared margin loss is

\[\tilde{\ell}_{\mu}(\theta;(X,Y)) = \begin{cases}
-\frac{2}{\mu}(\frac{1}{\mu} - 1)\log(1-\mu Y \cdot \hat{Y}) - \frac{2}{\mu} Y \cdot  \hat{Y} + C,\quad & \text{if } Y \cdot \hat{Y} \leq 0;\\
-\frac{2}{\mu}(\frac{1}{\mu} + 1)\log(1+\mu Y \cdot \hat{Y}) + \frac{2}{\mu} Y \cdot  \hat{Y} + C,\quad & \text{if } Y \cdot \hat{Y} \in (0, 1);\\
0, \quad & \text{if } Y \cdot \hat{Y} \geq 1,
\end{cases}\]
where the prediction $\hat{Y} = \theta^\top X$, the constant $C = \frac{2}{\mu}(\frac{1}{\mu} + 1)\log(1+\mu) - \frac{2}{\mu}$, and the hyper-parameter $\mu>0$ is the same one that defines the margin-based uncertainty function.
\end{example}

\begin{example}[Continued] 
The equivalent loss for the threshold-based uncertainty and the logistic loss function is given by the following: 
\[\tilde{\ell}_{\gamma}(\theta;(X,Y)) = \begin{cases}
\log(1+\exp(\gamma)),\quad & \text{if } Y \cdot  \hat{Y} \leq -\gamma,\\
\log(1+\exp(-Y \cdot  \hat{Y})),\quad & \text{if } Y \cdot  \hat{Y} \in (-\gamma, \gamma),\\
\log(1+\exp(-\gamma)), \quad & \text{if } Y \cdot  \hat{Y} \geq \gamma
\end{cases}\]
where the prediction $\hat{Y} = \theta^\top X$ and the hyper-parameter $\gamma$ is the same one that specifies the threshold-based uncertainty function. 
\end{example}

These formulas follow by integrating \eqref{eq:key_ODE}; Appendix~\ref{apd:equiv_loss} gives the calculations. With the fixed-step update, they identify the population objective followed by uncertainty sampling. Its calibration and curvature can now be studied through an explicit loss rather than through the querying protocol alone.

\subsection{Surrogate Property of Equivalent Loss}
\label{subsec:certify_uncert}

The derivation of equivalent loss makes it clear the objective function of the uncertainty sampling procedure. Following the principles of \citet{bartlett2006convexity}, we can examine the suitability of an equivalent loss and hence certify the properness of the uncertainty function. Recall that the minimax optimal link function $\psi$'s derivation is given in \citet{bartlett2006convexity} (also, see our Appendix \ref{apd:surrogate}). In the following proposition, we re-examine the previous examples and calculate the corresponding link functions against binary loss.

The probabilistic example makes the statistical meaning concrete. Fix a feature value, let $p=\Pr(Y=+1\mid X)$, and regard the predicted probability $q$ as a free scalar. Because $U(q)$ is independent of the unobserved label, the derivative of its conditional equivalent risk is
\[
\frac{\mathrm d}{\mathrm dq}\E[\tilde\ell(q,Y)\mid X]
=U(q)\left(-\frac pq+\frac{1-p}{1-q}\right)
=U(q)\frac{q-p}{q(1-q)}.
\]
For positive $U$, its sign is the sign of $q-p$, so the conditional optimum remains $q=p$. This pointwise conclusion and the parameterized optimization problem are distinct: a restricted model class can couple predictions across different features, and reweighting those features can change the best common parameter. The equivalent loss makes both the preserved conditional target and the altered population compromise visible.

\begin{figure}[!htbp]
\centering
\includegraphics[width=.72\textwidth]{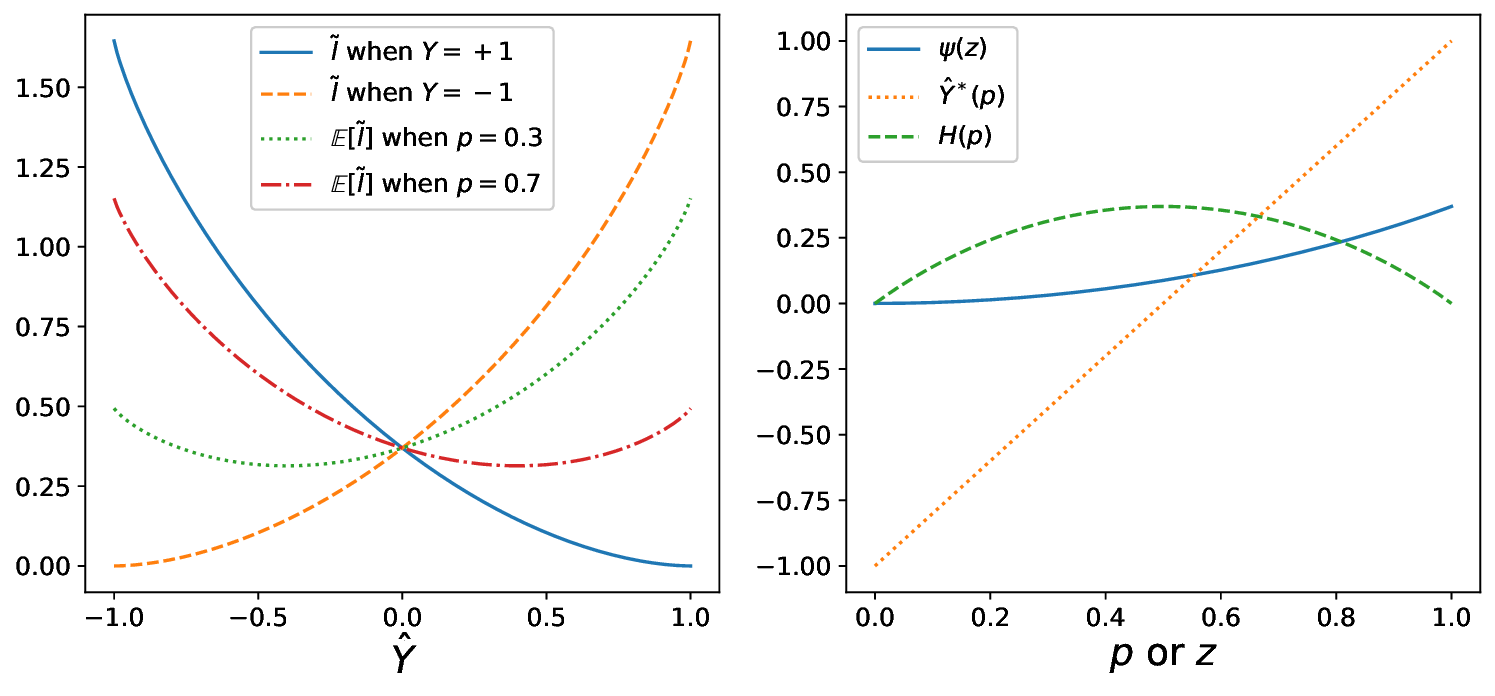}
\caption{Probabilistic model with entropy uncertainty (Example 1). The left subfigure shows $\tilde{\ell}(\hat{Y}, +1)$, $\tilde{\ell}(\hat{Y}, -1)$, and two different expected losses for positive probability $p = 0.3$ and $p = 0.7$. The minima of these expectations are the values $H(p)$, and the minimizing arguments are the values $\hat{Y}^*(p)$. The right subfigure shows $H(p)$ and $\hat{Y}^*$ as a function of $p$, and the surrogate link function $\psi$-transform $\psi(z)$.}
\label{fig:entropy}
\end{figure}

\begin{figure}[!htbp]
\centering
\includegraphics[width=.72\textwidth]{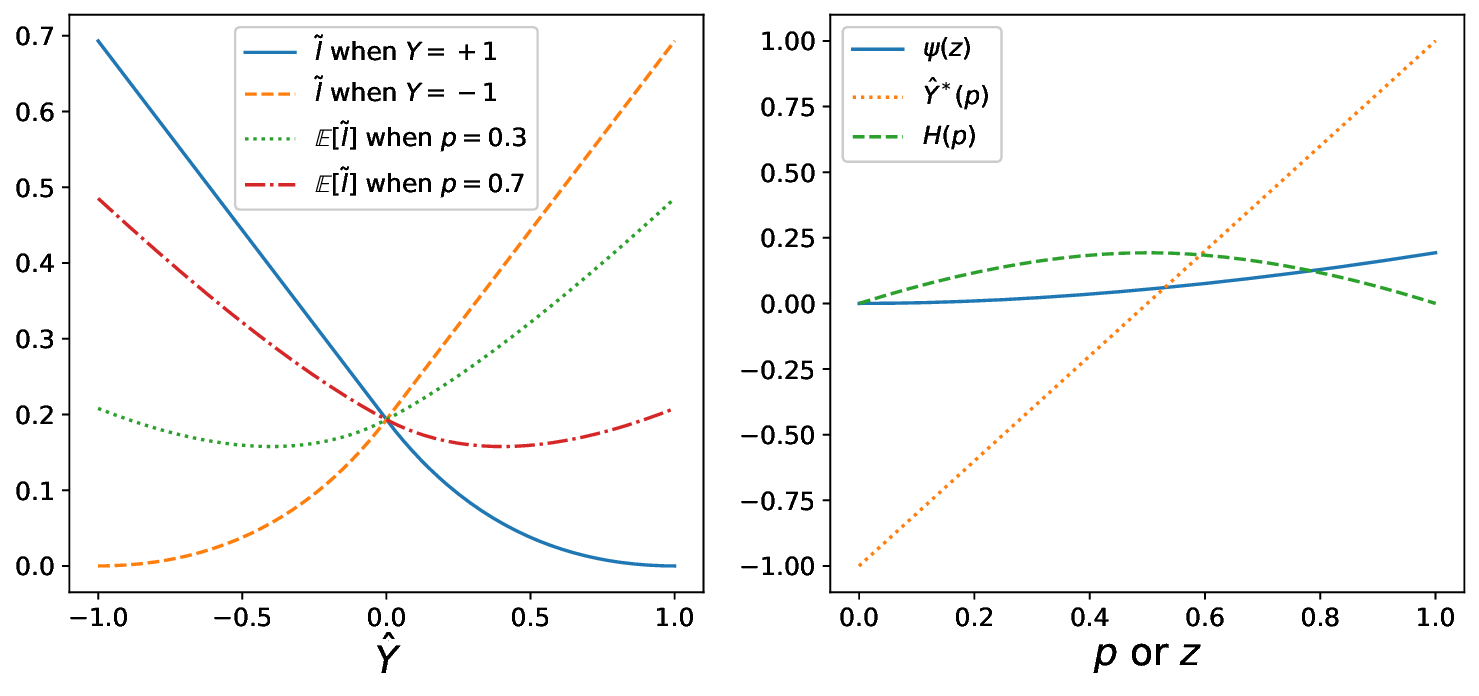}
\caption{Probabilistic model with least confidence uncertainty \citep{culotta2005reducing}.}
\label{fig:leastconfidence}
\end{figure}

\begin{figure}[!htbp]
\centering
\includegraphics[width=.72\textwidth]{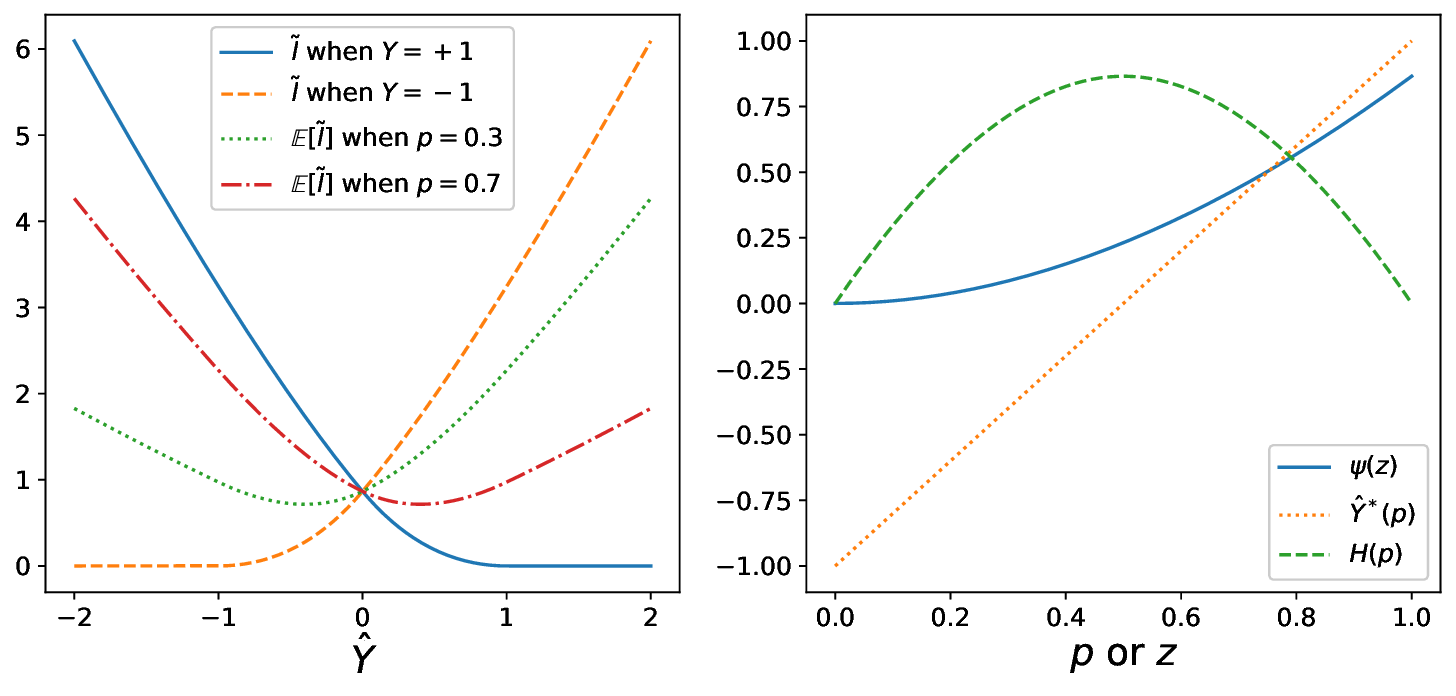}
\caption{Margin-based model \citep{raj2022convergence} with $\mu = 0.5$.}
\label{fig:raj}
\end{figure}

\begin{figure}[!htbp]
\centering
\includegraphics[width=.72\textwidth]{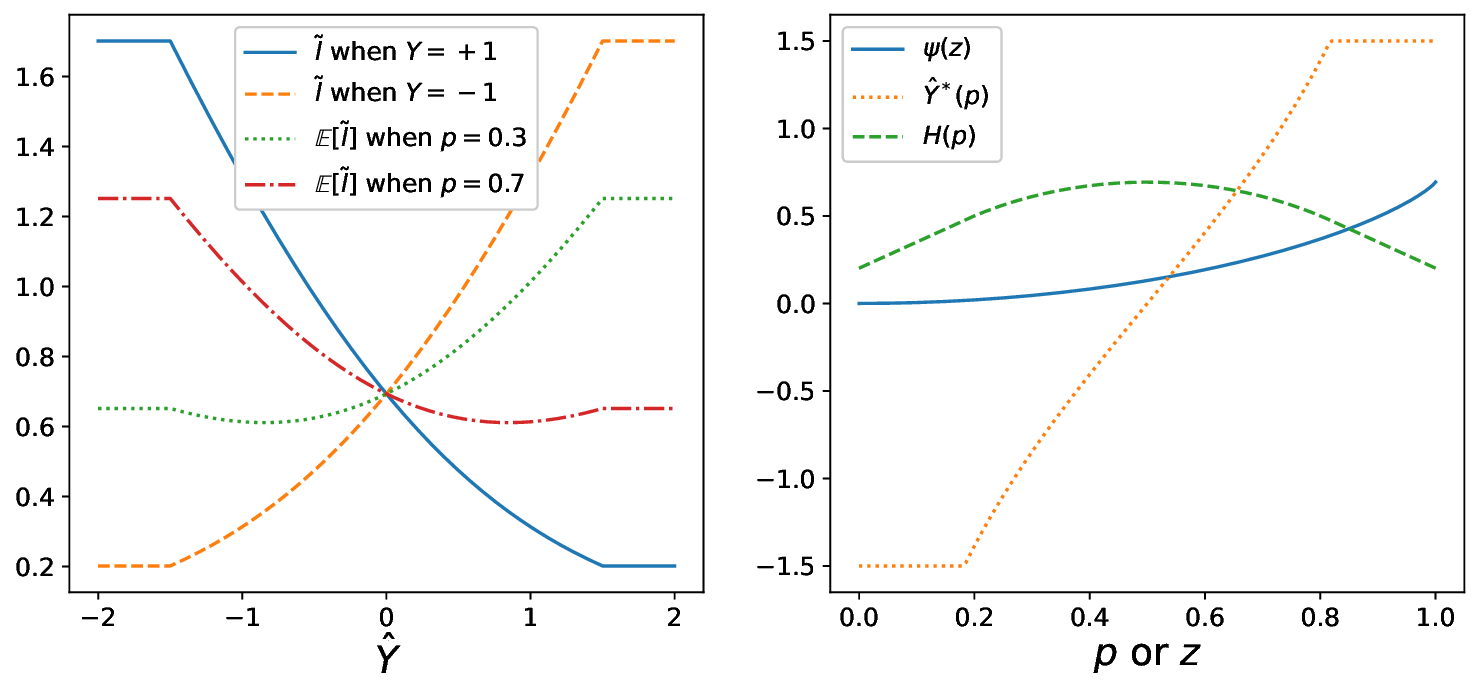}
\caption{Threshold-based model \citep{tifrea2022uniform} with $\gamma = 1.5$.}
\label{fig:tifrea}
\end{figure}

\begin{proposition}
\label{prop:link_calculate}
The equivalent losses in Examples~1--3 are classification-calibrated surrogate losses. Their optimal link functions have the following local behavior as $z\to0$:
\[
\begin{array}{c|c}
\text{uncertainty rule}&\psi(z)\\ \hline
\text{entropy}&\frac{\log 2}{2}z^2+o(z^2)\\
\text{least confidence}&\frac14z^2+o(z^2)\\
\text{smooth margin}&z^2+o(z^2)\\
\text{threshold}&\frac12z^2+o(z^2).
\end{array}
\]
The exact formulas and conditional-risk calculations are given in Appendix~\ref{apd:surrogate}; Figures~\ref{fig:entropy}--\ref{fig:tifrea} illustrate their geometry.
\end{proposition}

As noted earlier, the link function helps to transfer the excess risk bound under the equivalent loss to that under the binary loss. In the next section, we pursue such a roadmap by first establishing the convergence rate under the equivalent loss and then transferring it to a performance guarantee under the binary loss.

\FloatBarrier

\section{Sample Complexity for Binary Classification}

The stream-based learner can inspect abundant unlabeled observations, but labels remain costly. Our goal is to express the excess binary risk of Algorithm~\ref{alg:USGD} in terms of its expected label usage and to compare the resulting bound with passive SGD, which labels every observation.

The equivalent loss separates this analysis into optimization and statistical calibration. We first use convexity to bound the excess equivalent risk, retaining the query rate in the gradient's second moment. The surrogate link then converts that bound to binary risk. Because convexity can be lost under uncertainty weighting, the examples that follow identify which loss--uncertainty pairs support this argument and which introduce an optimization obstruction. Finally, a matched comparison shows how the average query rate, maximal uncertainty, and fixed learning-rate scale enter the leading classification bound.

\subsection{Convergence Analysis for Convex Loss}
\label{subsec:convergence}

For a fixed learning rate, the population drift in \eqref{eq:history-gradient} is the gradient of the expected equivalent loss. The remaining ingredient is the variance of the queried direction: observations with small query probability contribute less to its second moment. Keeping this contribution explicit yields a useful family of bounds indexed by a single tuning constant.

\begin{definition}[Loss convexity]
A loss $\ell(\theta;(X,Y))$ is convex if it is convex in $\theta$ for every $(X,Y)\in\mathcal X\times\mathcal Y$.
\end{definition}

Let $\tilde\theta^*$ minimize $\E[\tilde\ell(\theta;(X,Y))]$ over $\Theta$, and retain the query-count notation
\[
Q_T=\sum_{t=1}^T U_t,\qquad r_T=\frac{Q_T}{T}.
\]
Although $Q_T$ and $r_T$ are random, their expectations describe the algorithm's label usage:
\[
\E[Q_T]=\E\!\left[\sum_{t=1}^T\mathbbm{1}\{\xi_t\le U_t\}\right]
=T\E[r_T].
\]
The risk bounds below involve these expectations, while the updates use neither the realized terminal count nor the future query rate.

\begin{proposition}[Fixed-step equivalent-risk bound]
\label{prop:SGD_converge}
Suppose all iterates lie in a convex set $\Theta$, the equivalent loss is convex, and $\|\theta_1-\tilde\theta^*\|_2\le D$. Assume differentiation and population expectation may be interchanged and
\begin{equation}
\label{eq:feature-second-moment}
\E\!\left[\|\nabla_\theta\ell(\theta_t;(X_t,Y_t))\|_2^2
\mid\mathcal F_t^X\right]\le G^2
\quad\text{almost surely.}
\end{equation}
For a deterministic $c>0$ selected before the stream, set
\[
\eta_t=\frac{cD}{G\sqrt T},\qquad
\bar\theta_T=\frac1T\sum_{t=1}^T\theta_t.
\]
Then Algorithm~\ref{alg:USGD} satisfies
\begin{align}
&\E[\tilde\ell(\bar\theta_T;(X,Y))]
-\E[\tilde\ell(\tilde\theta^*;(X,Y))]\notag\\
&\qquad\le \frac{DG}{2\sqrt T}
\left(\frac1c+c\E[r_T]\right).
\label{eq:constant-c-bound}
\end{align}
The test pair $(X,Y)\sim\mathcal P$ is independent of training, and expectations include the algorithm's randomness.
\end{proposition}

Condition~\eqref{eq:feature-second-moment} holds, for example, when the original sample gradients are uniformly bounded by $G$. Conditioning on the feature is useful because the query decision also depends on it. In particular,
\[
\E[\|g_t\|_2^2\mid\mathcal F_t]
\le G^2\E[U_t\mid\mathcal F_t].
\]
The squared-distance argument then produces two contributions: $DG/(2c\sqrt T)$ from the initial distance and $cDG\E[r_T]/(2\sqrt T)$ from queried-gradient variance. A larger $c$ reduces the first term but increases the second. The original weighted average in Algorithm~\ref{alg:USGD} becomes the ordinary average above when all $\eta_t$ are equal, so its normalizer is deterministic as well.

\begin{remark}[Choosing the fixed step size]
\label{rem:choosing-c}
Suppose $U_t\le U_{\text{max}}\le1$. Without an estimate of the average rate, the feasible choice $c=U_{\text{max}}^{-1/2}$ balances the worst-case terms and gives
\begin{equation}
\label{eq:umax-step-bound}
\frac{DG\sqrt{U_{\text{max}}}}{2\sqrt T}
\left(1+\frac{\E[r_T]}{U_{\text{max}}}\right)
\end{equation}
as an upper bound in \eqref{eq:constant-c-bound}. If an oracle could calibrate the constant so that $c^{-2}=\E[r_T]$, the two terms would instead balance at
\begin{equation}
\label{eq:oracle-step-bound}
\frac{DG\sqrt{\E[r_T]}}{\sqrt T}
=\frac{DG\E[r_T]}{\sqrt{\E[Q_T]}}.
\end{equation}
The rate belongs to the run under the chosen $c$, making this a self-consistent oracle balance. Average uncertainty from an earlier window or a pilot run can guide $c$ for a subsequent run. This suggests calibrating to typical uncertainty, with $U_{\text{max}}$ as a feasible reference. Appendix~\ref{apd:step-calibration} quantifies sensitivity to an imperfect fixed estimate.
\end{remark}

The next theorem places the optimization bound on the binary-risk scale through the equivalent-loss link, retaining approximation error for the chosen hypothesis class. This joint treatment matters because rescaling $U$ changes both the equivalent loss and its link.

\begin{theorem}[Excess binary risk]
\label{thm:stream-based}
Under Proposition~\ref{prop:SGD_converge}, suppose $\tilde\ell$ admits a convex, nondecreasing surrogate link $\psi$. Then
\begin{align}
&\E[L_{\mathrm{01}}(f_{\bar\theta_T})]-\inf_{g\in\mathcal G}L_{\mathrm{01}}(g)\notag\\
&\le\psi^{-1}\!\left(
\frac{DG}{2\sqrt T}\left(\frac1c+c\E[r_T]\right)
+\underbrace{\left(
\inf_{\theta\in\Theta}\E[\tilde\ell(f_\theta(X),Y)]
-\inf_{g\in\mathcal G}\E[\tilde\ell(g(X),Y)]
\right)}_{\text{approximation error of }\Theta}
\right).
\label{eq:stream-risk-transfer}
\end{align}
Here $\psi^{-1}$ denotes the nondecreasing generalized inverse, and $\mathcal G$ is the class of all measurable predictors.
\end{theorem}

The first term measures optimization and sampling error; the second measures the expressive gap between the prescribed model class and unrestricted prediction. These terms describe different obstacles. A richer class may reduce approximation error without improving the geometry of its equivalent loss. Conversely, convexity can make optimization tractable even when the model remains misspecified. When approximation error vanishes, the link's local behavior determines how quickly optimization accuracy becomes classification accuracy. For the examples above, $\psi(z)=\Theta(z^2)$ near zero, so the transfer takes a square root. The following convexity analysis identifies where the optimization bound applies, after which we compare the active and passive bounds at a common label budget.

\subsection{Convexity under Equivalent Loss}
\label{subsec:convexity}
Calibration and convexity answer different questions. For a twice differentiable margin loss $\ell(s)$ and uncertainty $U(s)$, where $s=Y\theta^\top X$,
\begin{equation}
\label{eq:curvature-decomposition}
\nabla_\theta^2\tilde\ell
=XX^\top\{U'(s)\ell'(s)+U(s)\ell''(s)\}.
\end{equation}
The second term inherits the original curvature; the first is created by selective querying. If $\ell$ is decreasing and $U$ decreases with $|s|$, this additional term has opposite signs on the two sides of the decision boundary. The negative-margin side is particularly delicate: a confidently incorrect observation receives a large loss gradient but a small query probability.

\setcounter{example}{0}
\begin{example}[Probabilistic uncertainty under logistic regression]
In Example~1, set $q=(1+\exp(-\theta^\top X))^{-1}$ and write $z=\theta^\top X$. Both entropy and least-confidence uncertainty can produce a non-convex equivalent loss in $z$, even though cross entropy is convex in this coordinate. For least confidence and $Y=1$, the negative-margin branch has derivative $-q(1-q)$ and curvature
\[
\tilde\ell''(z)=-q(1-q)(1-2q)<0,\qquad z<0.
\]
The equivalent loss remains convex in the probability coordinate $q$; the nonlinear logistic parameterization changes the geometry seen by the optimizer. The entropy calculation is in Appendix~\ref{apd:convex-details}.
\end{example}

\begin{example}[Squared margin loss]
For $U_\mu(s)=(1+\mu|s|)^{-1}$ and $\ell(s)=(1-s)_+^2$, the equivalent loss in Example~2 has curvature
\[
\tilde\ell_\mu''(s)=
\begin{cases}
2(1-\mu)/(1-\mu s)^2,&s<0,\\
2(1+\mu)/(1+\mu s)^2,&0<s<1,\\
0,&s>1.
\end{cases}
\]
The first derivative is continuous at the junctions, so convexity is retained for $\mu\in[0,1]$. This example demonstrates that a nonconstant uncertainty score can preserve an optimizable equivalent objective.
\end{example}

\begin{example}[Threshold uncertainty]
The loss in Example~3 is logistic inside the queried margin band and constant outside it. At the left edge of the band, the derivative jumps downward from zero to a negative value. Hence the equivalent loss is non-convex. The flat tails make clear why correct classification calibration can coexist with unfavorable stationary behavior; see Figure~\ref{fig:tifrea}.
\end{example}

\newcounter{hingecounter}\setcounter{hingecounter}{\value{example}}
\begin{example}[Ordinary margin loss]\label{eg:Hinge}
For $\ell(s)=(1-s)_+$ and $U_\mu(s)=(1+\mu|s|)^{-1}$, direct integration produces a piecewise logarithmic equivalent loss with link
\[
\psi_\mu(z)=\frac{\log(1+\mu)}{\mu}z.
\]
Thus the attractive linear surrogate link survives, but the negative-margin branch has negative curvature whenever $\mu>0$. The loss formula is given in Appendix~\ref{apd:convex-details} and plotted in Figure~\ref{fig:hinge}.
\end{example}

\begin{proposition}\label{prop:Hinge_non-conv}
For the margin loss $\ell(s)=(1-s)_+$, consider $U(s)=h(|s|)$ with $h$ positive, non-increasing, and piecewise differentiable. If the equivalent loss is continuous and convex as a scalar margin loss for both labels, then $h$ is constant. In particular, a nontrivial decreasing margin-based uncertainty cannot preserve global convexity for the ordinary margin loss.
\end{proposition}

The proposition identifies the compatibility condition behind \eqref{eq:curvature-decomposition}: curvature supplied by the original loss must offset variation in the query score. The ordinary margin loss has zero curvature away from its kink, while its squared version supplies enough curvature for a nontrivial range of $\mu$. This distinction explains why changing the training loss can fundamentally change the behavior of the same querying rule.

\begin{figure}[!htbp]
\centering
\includegraphics[width=.72\textwidth]{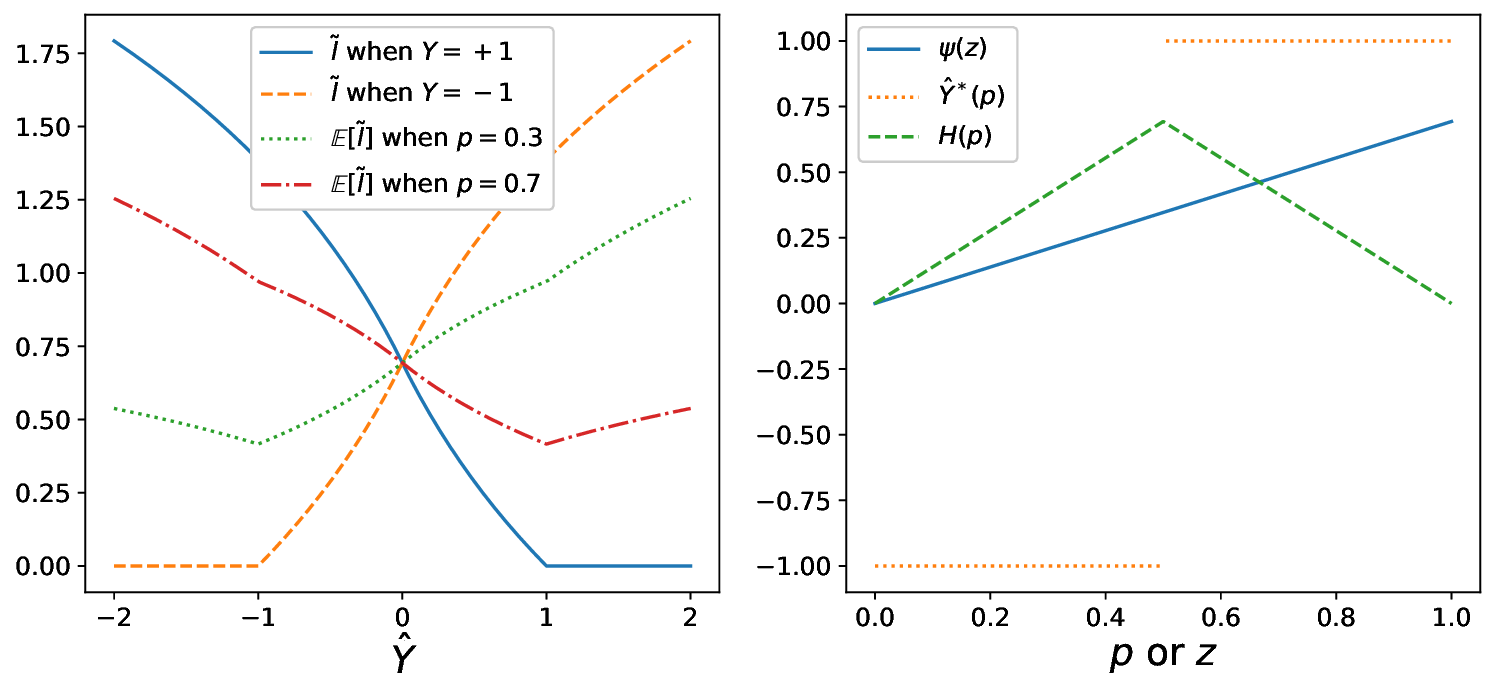}
\caption{Margin loss with the margin-based uncertainty when $\mu = 0.5$.}
\label{fig:hinge}
\end{figure}

\begin{example}[Exponential loss]\label{eg:exponential}
Let $\ell(s)=\exp(-s)$ and $U_\mu(s)=\exp(-\mu|s|)$ for $\mu\in[0,1)$. The equivalent loss is
\[
\tilde\ell_\mu(s)=
\begin{cases}
(1+\mu)^{-1}\exp(-(1+\mu)s)+\mu/(1+\mu),&s\ge0,\\
(1-\mu)^{-1}\exp(-(1-\mu)s)-\mu/(1-\mu),&s<0.
\end{cases}
\]
It remains convex and has $\psi_\mu(z)=z^2/2+o(z^2)$ near zero. Scalar curvature is bounded below on compact margin intervals; parameter-level strong convexity additionally requires nondegenerate feature geometry. Together with the squared-margin example, this supplies a positive counterpart to the threshold and ordinary-margin obstructions.
\end{example}

\begin{figure}[!htbp]
\centering
\includegraphics[width=.72\textwidth]{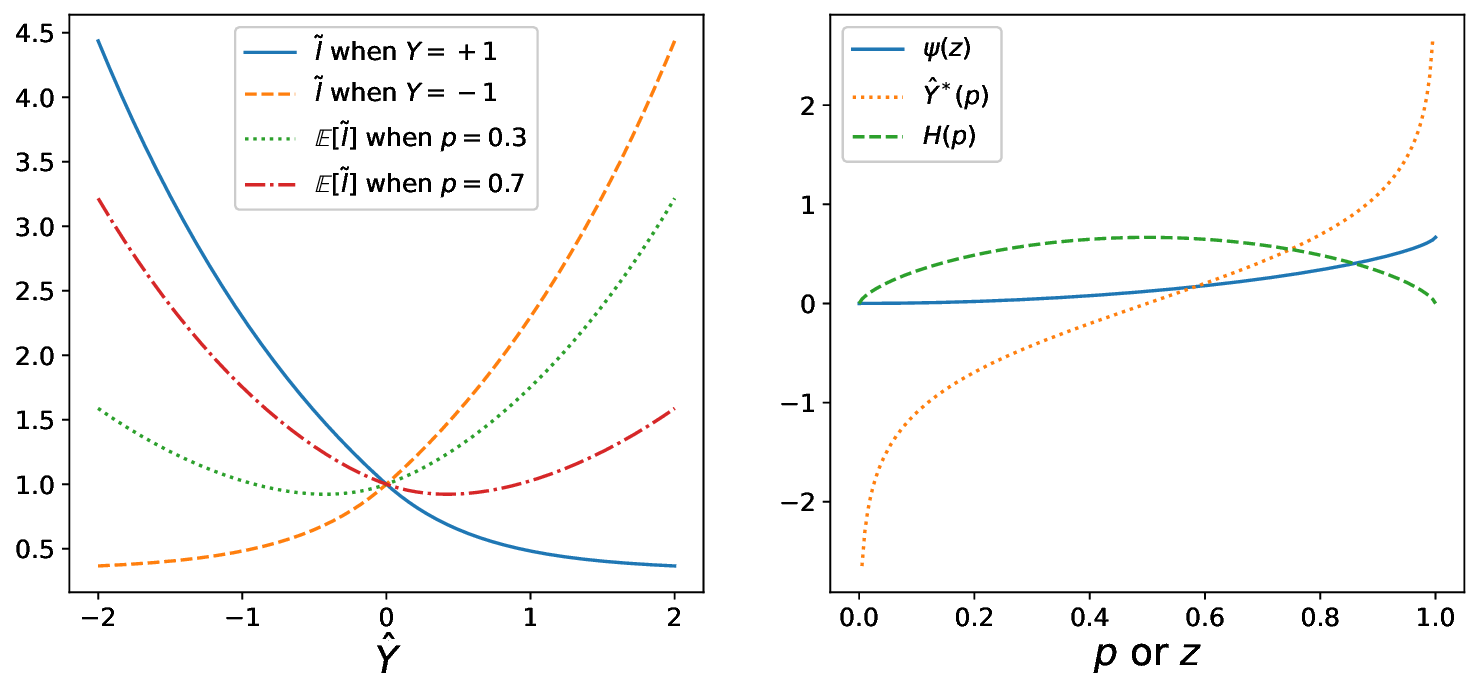}
\caption{Exponential loss and exponential uncertainty with $\mu = 0.5$.}
\label{fig:exponential}
\end{figure}

\FloatBarrier

\subsection{Sample Complexity via Link Function}
\label{subsec:link_func}
Uncertainty weighting changes both curvature and calibration. Once the equivalent loss remains convex, Proposition~\ref{prop:SGD_converge} supplies the optimization bound. The local link calculation below identifies how its query-rate factor enters a comparison with passive learning at the same expected number of labels.

We start from the following proposition, stating that the surrogate link function $\psi(z)$ of the original loss asymptotically behaves the same as $\tilde{\psi}(z)$ of the equivalent loss under some mild conditions. The term ``asymptotically'' means that we are considering cases when the excess risk approaches zero ($z$ approaches zero).

\begin{assumption}
\label{assm:uncertainty}
Assume that $U$ is an even and twice continuously differentiable function of $\theta^\top X$ and takes maximum at $U(0) \eqqcolon U_{\text{max}}$. Also, we assume without loss of generality that $U > 0$.
\end{assumption}
Assumption~\ref{assm:uncertainty} describes smooth, symmetric uncertainty scores that are largest at the decision boundary.

\begin{assumption}
\label{assm:loss_minimizer}
Assume that $\ell$ is a decreasing and twice continuously differentiable function of $Y \cdot \theta^\top X$ (denoted by $s$ for notation simplicity).  Following \citet{bartlett2006convexity}, we denote the probability of a positive $Y$ by $p$, and we denote the expected loss $\ell$ and the expected equivalent loss $\tilde{\ell}$ as
\[
C_p(s) \coloneqq p \ell(s) + (1-p) \ell(-s).
\]
We assume that for any $p \in (0, 1)$, there is a unique minimizer of $C_p$, denoted by $s^\ast(p)$. We assume that $C_p(s)$ is convex on $\mathbb R$ for every $p\in(0,1)$. Furthermore, we assume that $s^\ast(p)$ is a twice differentiable function of $p$ for $p \in (0, 1)$.
\end{assumption}
These conditions concern the original conditional risk. Logistic and exponential losses satisfy them; the direct calculations for squared-margin loss give the same local expansion despite its piecewise smoothness. Appendix~\ref{apd:surrogate} also treats the probabilistic and threshold examples directly.

Now, equipped with those assumptions, we are ready to present our main result in this subsection.

\begin{proposition}
\label{prop:equiv_surrogate}
Assume the equivalent loss $\tilde{\ell}$ exists for $s = Y \cdot \theta^\top X$ such that $U \cdot \frac{\partial \ell}{\partial s} = \frac{\partial \tilde{\ell}}{\partial s}$. Under Assumptions \ref{assm:uncertainty} and \ref{assm:loss_minimizer}, we have for $z$ near zero,
\[
\psi(z) = -\frac18 \cdot \frac{\mathrm{d}^2 H}{\mathrm{d} p^2}\bigg|_{p=\frac12} \cdot z^2 + o(z^2),
\]
and
\[\tilde{\psi}(z) = -U_{\text{max}} \cdot\frac18 \cdot \frac{\mathrm{d}^2 H}{\mathrm{d} p^2}\bigg|_{p=\frac12} \cdot z^2 + o(z^2) = U_{\text{max}} \cdot \psi(z) + o(z^2).\]
\end{proposition}

The curvature coefficient in this expansion is shared by the original and equivalent losses, up to $U_{\text{max}}$. We now combine it with the fixed-step bound.

\begin{corollary}[Matched expected-label upper bounds]
\label{coro:final_sample_complexity}
Suppose the conditions of Propositions~\ref{prop:SGD_converge} and~\ref{prop:equiv_surrogate} hold, the equivalent-loss approximation error in Theorem~\ref{thm:stream-based} is zero, and $-H''(1/2)>0$. If
\[
\frac{c^{-1}+c\E[r_T]}{\sqrt T}\longrightarrow0,
\]
then
\begin{align}
&\E[L_{\mathrm{01}}(f_{\bar\theta_T})]-\inf_{g\in\mathcal G}L_{\mathrm{01}}(g)\notag\\
&\quad\le
\left(
\frac{4DG}{U_{\text{max}}[-H''(1/2)]\sqrt T}
\left(\frac1c+c\E[r_T]\right)
\right)^{1/2}(1+o(1)).
\label{eq:fixed-c-classification}
\end{align}
Suppose also that passive SGD on the original convex loss has zero approximation error and the same valid bounds $D,G$. At an expected label budget $\E[Q_T]\to\infty$, its corresponding leading bound is
\[
\left(\frac{8DG}{[-H''(1/2)]\sqrt{\E[Q_T]}}\right)^{1/2}.
\]
The ratio of the leading active and passive bounds is therefore
\begin{equation}
\label{eq:fixed-c-risk-ratio}
\left[
\frac{\sqrt{\E[r_T]}}{2U_{\text{max}}}
\left(\frac1c+c\E[r_T]\right)
\right]^{1/2}.
\end{equation}
\end{corollary}

For the feasible choice $c=U_{\text{max}}^{-1/2}$, this ratio becomes
\begin{equation}
\label{eq:feasible-risk-ratio}
\left(\frac{\E[r_T]}{U_{\text{max}}}\right)^{1/4}
\left[\frac12\left(1+\frac{\E[r_T]}{U_{\text{max}}}\right)\right]^{1/2}
\le1,
\end{equation}
with strict inequality whenever $\E[r_T]<U_{\text{max}}$. Thus, even without knowing the average query rate, a step-size calibration based on maximal uncertainty gives a leading upper bound at least as favorable as its passive counterpart at the same expected label budget. This comparison partly indicates a mechanism for label efficiency: observations well away from the decision boundary lower the average query rate, while calibration near the boundary is governed by $U_{\text{max}}$.

Under the oracle balance in Remark~\ref{rem:choosing-c}, the ratio in \eqref{eq:fixed-c-risk-ratio} improves to
\[
\left(\frac{\E[r_T]}{U_{\text{max}}}\right)^{1/2}.
\]
Both comparisons concern leading risk upper bounds at matched label usage. The factor $(1+\E[r_T]/U_{\text{max}})/2$ in \eqref{eq:umax-step-bound} first enters the surrogate-risk constant; conversion from the raw horizon $T$ to $\E[Q_T]$, followed by the quadratic surrogate link, accounts for the additional powers in \eqref{eq:feasible-risk-ratio}. The statements use a common gradient bound and initial-distance bound for the two objectives and the well-specified regime in which approximation error vanishes.

\subsection{Numerical Illustration}
We illustrate the equivalent-loss mechanism using the two-dimensional Gaussian-mixture construction of \citet{mussmann2018uncertainty}. Four components have coordinate standard deviation $0.5$, centers $(-2,0)$, $(2,0)$, $(0,-2)$, and $(0,2)$, and mixture weights $0.2$, $0.3$, $0.4$, and $0.1$. The first two components carry positive labels and the other two negative labels. We compare empirical minimization of the original loss, minimization of the equivalent loss, and uncertainty sampling from the same initial parameter. The uncertainty-sampling runs use a constant step size $10^{-4}$ for $10^7$ raw observations, so the query-conditional representation also identifies their population objective.

\begin{figure}[!htbp]
\centering
\begin{subfigure}[t]{.455\textwidth}
\includegraphics[width=\linewidth]{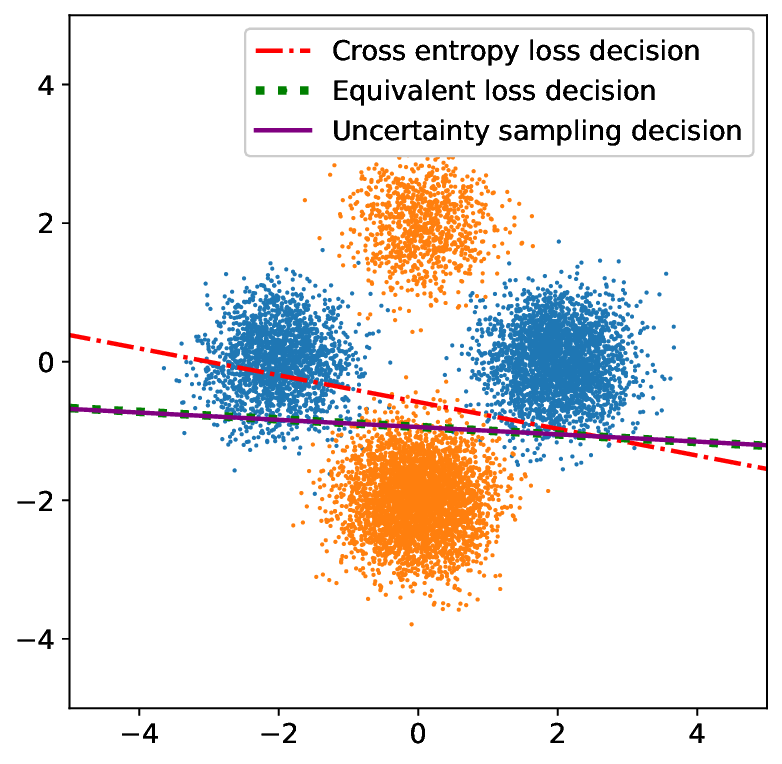}
\caption{Entropy uncertainty.}\label{fig:exp1-1-1}
\end{subfigure}\hfill
\begin{subfigure}[t]{.455\textwidth}
\includegraphics[width=\linewidth]{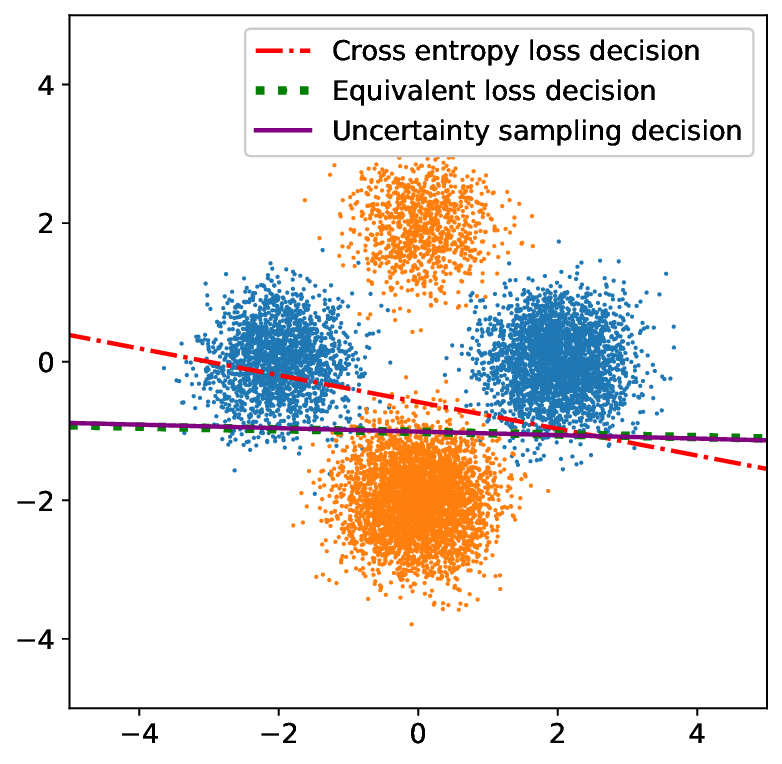}
\caption{Least-confidence uncertainty.}\label{fig:exp1-1-2}
\end{subfigure}
\caption{Probabilistic uncertainty: original-loss, equivalent-loss, and uncertainty-sampling decision boundaries.}
\label{fig:equivalent_loss}
\end{figure}

Figure~\ref{fig:equivalent_loss} shows that uncertainty sampling follows the equivalent-loss boundary rather than the boundary obtained from the original loss. The agreement illustrates the significance of the gradient-field calculation: selective acquisition does not simply supply a smaller training sample to the same objective. The induced objective changes, even when the original data distribution is unchanged.

\begin{figure}[!htbp]
    \centering
    
    \begin{subfigure}{0.475\textwidth}
        \centering
        \includegraphics[width=\textwidth]{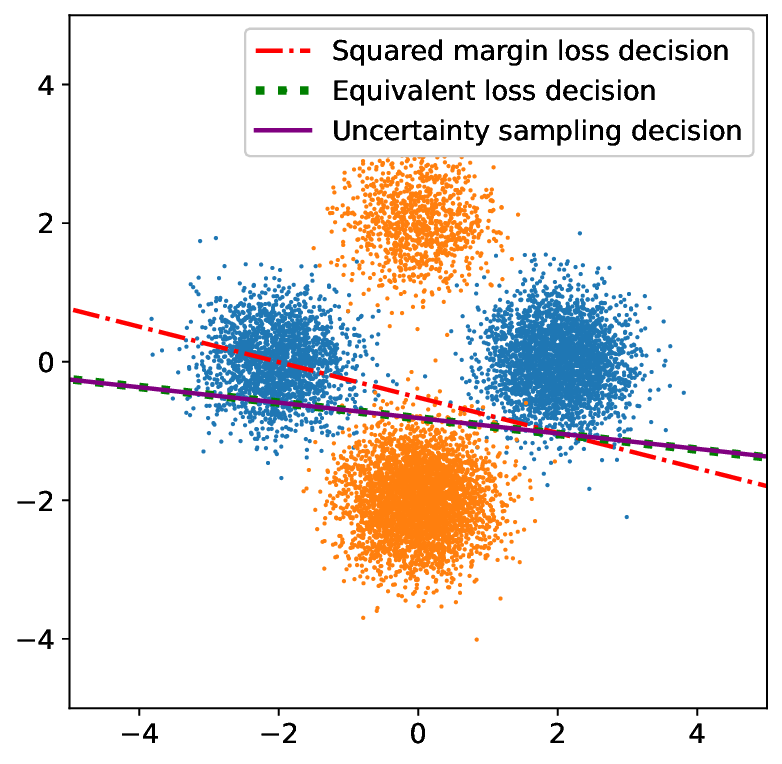}
        \caption{Example 2: squared margin loss with margin-based uncertainty ($\mu = 1$).}
        \label{fig:exp1-2}\label{fig:main-squared}
    \end{subfigure}
    \hfill
    \begin{subfigure}{0.475\textwidth}
        \centering
        \includegraphics[width=\textwidth]{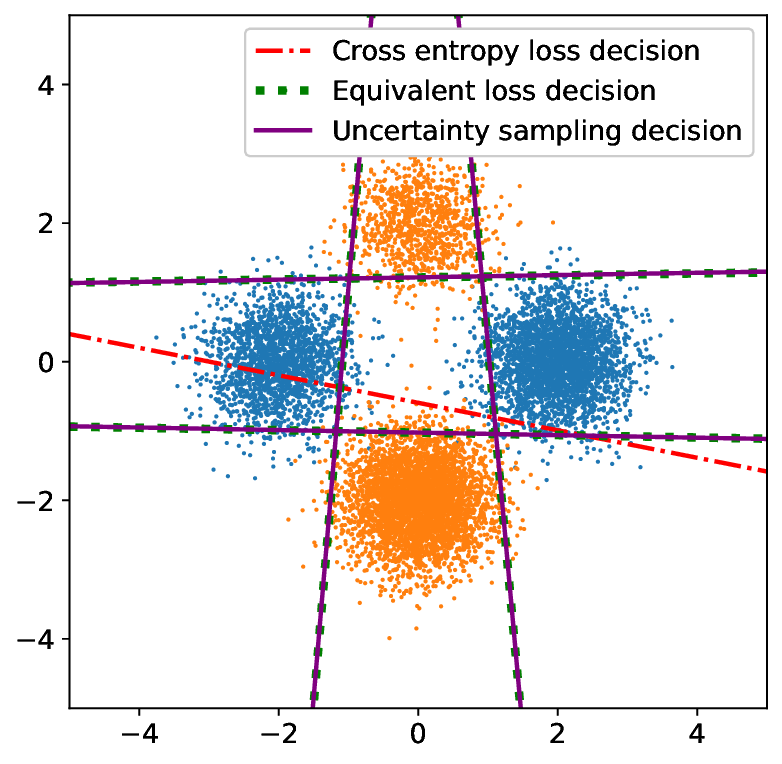}
        \caption{Example 3: logistic loss with threshold-based uncertainty ($\gamma = 2$).}
        \label{fig:exp1-3}\label{fig:main-threshold}
    \end{subfigure}

    \vspace{0.5em}

    \begin{subfigure}{0.475\textwidth}
        \centering
        \includegraphics[width=\textwidth]{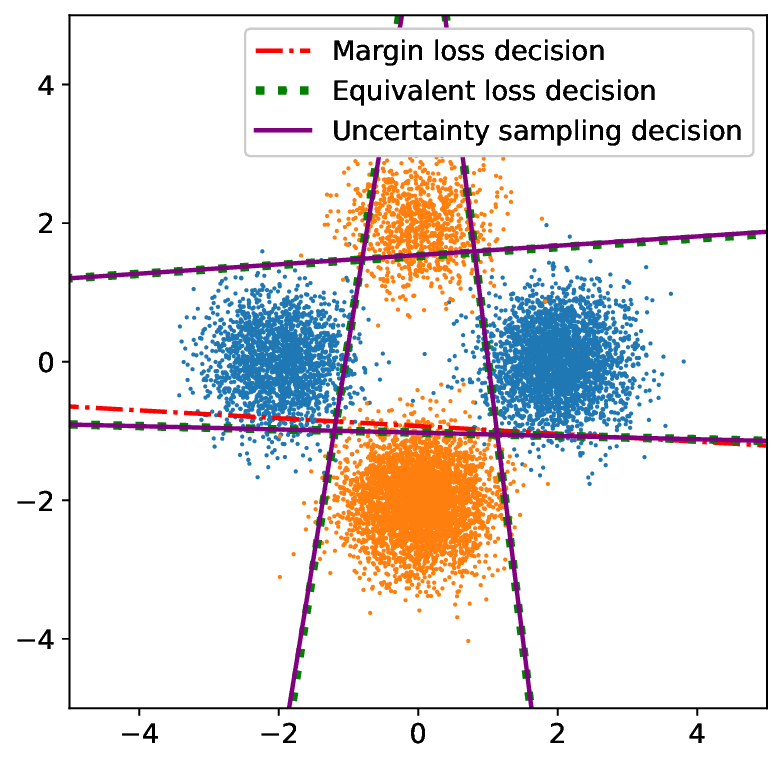}
        \caption{Example 4: margin loss with margin-based uncertainty ($\mu = 10$).}
        \label{fig:exp1-4}
    \end{subfigure}
    \hfill
    \begin{subfigure}{0.475\textwidth}
        \centering
        \includegraphics[width=\textwidth]{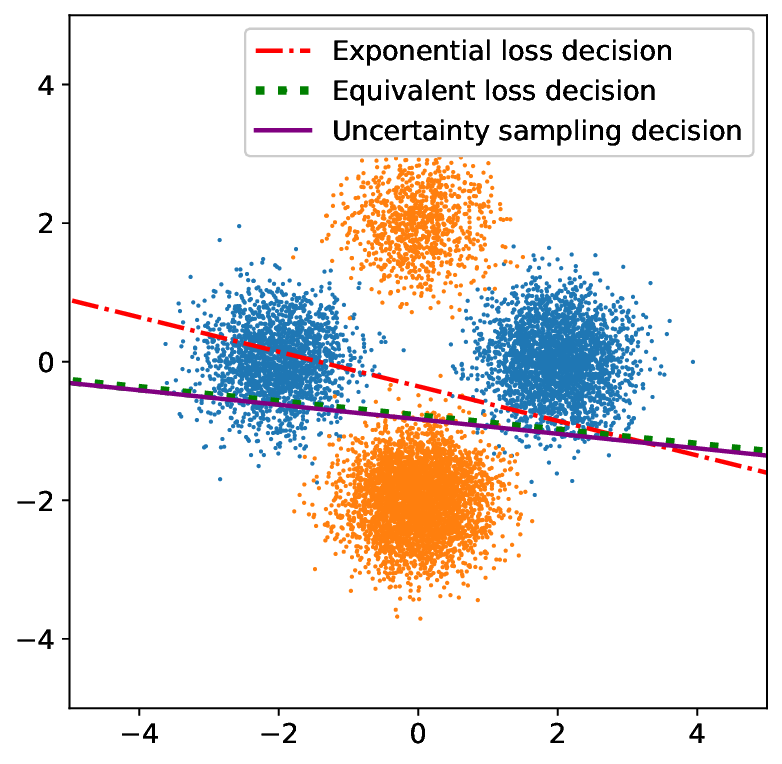}
        \caption{Example 5: exponential loss with exponential uncertainty ($\mu = 0.9$).}
        \label{fig:exp1-5}
    \end{subfigure}

    \caption{Decision boundaries under different loss functions and uncertainty measures across several examples.}
    \label{fig:exp1-all}\label{fig:main-geometry}
\end{figure}

Figure~\ref{fig:main-geometry} contrasts the squared-margin and threshold constructions. The former retains convexity over the parameter range discussed above; the latter introduces flat tails and non-convexity. This is why the framework provides more than a rate calculation: the equivalent loss predicts which geometry the algorithm encounters. The remaining panels show the ordinary-margin and exponential examples, completing the comparison across the loss--uncertainty pairs considered above.

\section{Pool-Based Setting}
\label{sec:pool-based}

In this section, we generalize our equivalent loss principle to the uncertainty sampling algorithm under the pool-based setting. Different from the stream-based setting, the features for all the samples are given at the beginning. To distinguish between the number of samples and the number of steps for the gradient descent algorithm, we use $i=1,\dots,n$ to index the samples and $t = 1,\dots T$ to index the gradient descent time steps.

Algorithm~\ref{alg:USGD_pool-based} presents the pool-based uncertainty sampling algorithm. At each time step, the algorithm calculates the uncertainty for each sample in the data pool $\mathcal{D}_n$ given the current model parameter $\theta_t.$ Then the algorithm samples an index according to the probability distribution proportional to the uncertainty and queries the label of the sampled index. Based on this new label, the algorithm updates the model parameter via gradient descent.

\begin{algorithm}[tb]
\caption{Uncertainty sampling with gradient descent update (pool-based version)}
    \label{alg:USGD_pool-based}
    \begin{algorithmic}[1] 
    \Require Unlabeled dataset $\mathcal{D}_n^X=\{X_i\}_{i=1}^{n}$, step size $\{\eta_t\}_{t=1}^T >0$, uncertainty function $U(\theta; X)$
    \State Initialize $\theta_1$; $\bar{\theta}_0 \leftarrow \theta_1$
    \For{$t=1,...,T$}
    \State Calculate the uncertainty $U(\theta_t; X_i)$ for each $X_i \in \mathcal{D}_n$ 
    \State Sample $i_t \in [n]$ according to the probability distribution $\propto U(\theta_t; X_i)$, where each sample has a probability of
    \[
    p_i = U(\theta_t; X_i) \Big/ \sum_{j=1}^n U(\theta_t; X_j)
    \]
    to be selected. Denote $\frac1n \sum_{j=1}^n U(\theta_t; X_j)$ as $S_t$
    \State Query a label $Y_t$ of $X_{i_t}$
    \State Update the parameter via gradient descent
    \[ \theta_{t+1} \leftarrow \theta_t - \eta_t\cdot  \frac{\partial \ell(\theta;(X_{i_t}, Y_t))}{\partial \theta}\Bigg\vert_{\theta=\theta_t} \]
    \State $\bar{\theta}_{t} \leftarrow (1-\frac{\eta_t/S_t}{\sum_{s=1}^t \eta_s/S_s}) \bar{\theta}_{t-1} + \frac{\eta_t/S_t}{\sum_{s=1}^t \eta_s/S_s} \theta_{t}$
    \EndFor
    \Ensure $\bar{\theta}_{T}$
    \end{algorithmic}
\end{algorithm}

\subsection{Theoretical Analysis}\label{subsec:pool_based_theory}
The pool makes the sampled feature endogenous: $\theta_t$ determines the uncertainty of every pool observation and therefore its probability of being selected. Conditional on the realized pool and optimizer history, however, the expected sampled gradient still has a simple form. Let
\[
S_t=\frac1n\sum_{i=1}^nU(\theta_t;X_i),
\qquad s_T=\frac1T\sum_{t=1}^T S_t.
\]
For fixed labels, write $\hat{\tilde\ell}(\theta)=n^{-1}\sum_i\tilde\ell(\theta;(X_i,Y_i))$. The sampled gradient $g_t$ obeys
\begin{equation}\label{eq:pool-mean}
\E[g_t\mid\mathcal F_t]=\frac1{S_t}\nabla_\theta\hat{\tilde\ell}(\theta_t),
\end{equation}
where the pool is included in $\mathcal F_t$. Unlike the current-feature stream step, $S_t$ is known before the random pool index is drawn. The output weights $\eta_t/S_t$ in Algorithm~\ref{alg:USGD_pool-based} reflect this normalization.

Population performance contains optimization, generalization, and approximation components. Writing $\hat\ell$ for empirical risk and $\ell$ for population risk as in the original notation, one can decompose
\begin{align*}
\E[\ell(\theta)]-\inf_{g\in\mathcal G}\E[\ell(g)]
&=\E[\ell(\theta)-\hat\ell(\theta)]
+\E[\hat\ell(\theta)]-\inf_{\theta'}\E[\hat\ell(\theta')]\\
&\quad+\inf_{\theta'}\E[\hat\ell(\theta')]-\E[\hat\ell(\theta^*)]\\
&\quad+\E[\hat\ell(\theta^*)-\ell(\theta^*)]
+\E[\ell(\theta^*)]-\inf_{g\in\mathcal G}\E[\ell(g)].
\end{align*}
The first and fourth terms are empirical-to-population discrepancies, the second is optimization error, the third is nonpositive, and the last is approximation error. The equivalent-loss representation acts on the optimization term; standard uniform-convergence tools then connect it to population risk.

The normalizer $S_t$ is known before the pool index is drawn, but the terminal sum of output weights is random. Its deterministic lower bound is enough for a valid optimization guarantee without modifying the sampling rule, update, or weighted output of Algorithm~\ref{alg:USGD_pool-based}.

\begin{proposition}[Pool optimization bound]
\label{prop:SGD_converge_pool}
Condition on the labeled pool (or on its conditional-label risk in the repeated-query model). Suppose $\hat{\tilde\ell}$ is convex, $\|\theta_1-\hat{\tilde\theta}^*\|_2\le D$, and the sampled gradient has conditional second moment at most $G^2$. Assume $0<S_t\le U_{\text{max}}$ throughout. For $\eta_t=D/(G\sqrt T)$ and the output weights in Algorithm~\ref{alg:USGD_pool-based},
\[
\E[\hat{\tilde\ell}(\bar\theta_T)-\hat{\tilde\ell}(\hat{\tilde\theta}^*)]
\le\frac{DGU_{\text{max}}}{\sqrt T}.
\]
\end{proposition}

The proof first bounds the unnormalized sum of empirical excess risks with weights $\eta_t/S_t$. Convexity and nonnegativity of these risks allow its random normalizer to be replaced by the pathwise lower bound $TD/(G\sqrt T\,U_{\text{max}})$. This retains the original pool algorithm while avoiding any factorization involving future weights. Generalization is handled separately: when the original loss is $\beta$-Lipschitz and $U\le U_{\text{max}}$, the equivalent loss is $U_{\text{max}}\beta$-Lipschitz. The usual Rademacher contraction bound can therefore be combined with the optimization and approximation terms. Additional details appear in Appendix~\ref{apd:pool-details}.

\subsection{Repeated-Query versus Single-Query}
A pool may attach one persistent label to each feature or permit repeated independent annotations. Under repeated querying, each visit to $X_i$ generates a fresh draw from $\mathcal P_{Y\mid X_i}$. Define
\[
L(\theta;X)=\E[\ell(\theta;(X,Y))\mid X],\qquad
\tilde L(\theta;X)=\E[\tilde\ell(\theta;(X,Y))\mid X].
\]
The gradient representation in \eqref{eq:pool-mean} then holds for the empirical conditional risk $n^{-1}\sum_i\tilde L(\theta;X_i)$. The acquisition mechanism is the same, while the statistical target averages annotation noise at each feature.

This distinction also appears in complexity calculations. For $m$ independent labels per feature, form the repeated-label loss class by averaging their losses. Introducing a uniformly selected annotation index at each feature and moving the supremum through the resulting expectation gives
\[
\E[\mathcal R_n(\mathcal L^{\mathrm{rep}})]
\le\E[\mathcal R_n(\mathcal L^{\mathrm{sin}})].
\]
Repeated annotation can therefore reduce the component of generalization error attributable to conditional label noise. Its value must be weighed against the opportunity to label more distinct features at a fixed total cost. Appendix~\ref{apd:pool-details} supplies the class definitions and the full argument.

\section{Extensions and Boundaries of the Equivalent-Loss Framework}
\label{sec:multi_dim}
The examples above share an effectively scalar prediction coordinate. Two extensions reveal different boundaries of the construction: vector-valued predictions create an integrability problem, while momentum makes the optimizer clock part of the dynamics.

\subsection{Existence Beyond Scalar Predictions}
If $U$ and $\ell$ depend on $\theta$ through a single scalar $s=s(X;\theta)$, direct integration gives
\[
\tilde\ell(s;Y)=\int_a^s U(t)\,\partial_t\ell(t;Y)\,\mathrm dt.
\]
For vector-valued predictions, $U\nabla_\theta\ell$ need not be conservative. The following equivalent conditions characterize precisely when it defines a potential.
\begin{proposition}
\label{prop:existence_equiv}
    Suppose for any $X$ and $Y$, $U(\theta; X)$ and $\ell(\theta; (X, Y))$ are both twice continuously differentiable functions of $\theta \in \Theta$. Assume $\Theta$ to be an open and simply connected subset of $\mathbb{R}^k$. Now we fix any $(X, Y)$ such that $U$ and $\ell$ can be regarded as functions of $\theta \in \Theta$. The following conditions are equivalent:
    \begin{enumerate}
        \item There exists a function $\tilde{\ell}$ satisfying the equivalent loss condition \eqref{eq:key_ODE}, i.e.,
        \[
        \nabla_\theta \tilde{\ell} = U \cdot \nabla_\theta \ell;
        \]
        \item For any $i, j \in [k]$,
        \[
        \partial_i U \cdot \partial_j \ell = \partial_j U \cdot \partial_i \ell.
        \]
        Here, we denote $\frac{\partial f}{\partial \theta_j}$ by $\partial_j f$ for any function $f$;
        \item For any point such that $\nabla \ell \neq 0$, we have
        \[
        \nabla_\theta U = \lambda(\theta) \cdot \nabla_\theta \ell,
        \]
        for some scalar field $\lambda(\theta)$;
        \item For any regular level set $L_c^{\text{reg}}$ such that $\lbrace \theta\in \Theta: \ell(\theta) = c, \nabla_\theta \ell \neq 0 \rbrace$, $U(S)$ is also a constant on any path-connected component of $L_c^{\text{reg}}$.
    \end{enumerate}
\end{proposition}

The coordinate condition removes every two-dimensional curl component. Geometrically, uncertainty may vary normal to a regular loss level set but not tangentially along it. Simply connectedness promotes this local property to a global potential; more generally, the associated differential form must be exact \citep{bott1982differential}. The construction $U=h(\ell)$ is sufficient, though stronger than necessary when loss level sets have several disconnected components.

A population-level construction is available after averaging over the label. For the conditional risk $L(\theta;X)=\E[\ell(\theta;(X,Y))\mid X]$, a score proportional to $L$ produces
\[
L(\theta;X)\nabla_\theta L(\theta;X)
=\nabla_\theta\left\{\frac12L(\theta;X)^2\right\}.
\]
A constant normalization to $[0,1]$ is needed to make this a query probability when $L$ is bounded. This construction trades an integration problem for an estimation problem: the learner must estimate the conditional loss at the individual feature, not merely an average calibration statistic. For squared-error regression, that conditional loss includes both predictive bias and conditional variance; it equals the variance when the prediction is the true conditional mean. The framework therefore identifies both a possible extension and the statistical quantity it would need to estimate.

\subsection{Momentum Methods and the Query Clock}
\label{subsec:momentum}
Consider a query-only momentum implementation: a rejected observation leaves the parameter, velocity, and optimizer counter fixed. Both uncertainty and gradient are evaluated at the same point; for Nesterov, this is the look-ahead point. Define the population query rate $r(\theta)=\E[U(\theta;X)]>0$.

\begin{proposition}[Embedded query chain]\label{prop:embedded-query-chain}
While the evaluation point is fixed at $\theta$, the next successfully queried observation has distribution
\[
\mathcal P^Q_\theta(\mathrm dx,\mathrm dy)
=\frac{U(\theta;x)}{r(\theta)}\mathcal P(\mathrm dx,\mathrm dy).
\]
Thus
\[
\E_{\mathcal P^Q_\theta}[\nabla_\theta\ell(\theta;(X,Y))]
=\frac1{r(\theta)}\nabla_\theta\E_{\mathcal P}[\tilde\ell(\theta;(X,Y))].
\]
\end{proposition}

The result is exact because the state is frozen during the geometrically distributed waiting time. It identifies the force seen by a momentum method indexed by successful queries. The following ODE refers to the usual common-step inertial scaling of that queried chain; the fixed-step first-order analysis in Proposition~\ref{prop:SGD_converge} concerns a different optimizer.

\begin{proposition}[Query-clock momentum ODE]\label{prop:momentum-ode}
Assume the normalized mean field is locally Lipschitz, queried gradients have locally bounded second moments, and the interpolated iterates remain in a compact subset of the parameter domain on the time interval of interest. Under the inertial scaling specified in Appendix~\ref{apd:momentum-details}, the interpolation converges in probability to
\begin{equation}\label{eq:momentum-ode}
\ddot\theta(s)+a(s)\dot\theta(s)
+\frac1{r(\theta(s))}\nabla_\theta\E[\tilde\ell(\theta(s);(X,Y))]=0.
\end{equation}
Constant $a(s)=\gamma>0$ gives heavy-ball dynamics; $a(s)=3/s$ gives the classical Nesterov equation on intervals bounded away from zero \citep{polyak1964some,su2016differential}. Constant solutions have $\dot\theta=0$ and zero gradient of the expected equivalent loss.
\end{proposition}

The positional stationary set is therefore unchanged by the normalization. The energy
\[
\mathcal E(s)=\E[\tilde\ell(\theta(s);(X,Y))]
+\tfrac12r(\theta(s))\|\dot\theta(s)\|_2^2
\]
satisfies
\begin{equation}\label{eq:momentum-energy}
\dot{\mathcal E}(s)
=-\left(a(s)r(\theta(s))-\tfrac12\frac{\mathrm d}{\mathrm ds}r(\theta(s))\right)
\|\dot\theta(s)\|_2^2.
\end{equation}
Unlike first-order flow, where division by $r$ merely changes time, inertia turns variation in the clock into a damping contribution. A falling query rate adds dissipation. At a stationary point, the normalized-field Jacobian is the Hessian of the expected equivalent loss divided by $r(\theta^*)$. For constant positive damping, a nondegenerate minimum is locally stable and a negative-curvature direction remains unstable; the query rate rescales local stiffness.

If instead every raw observation advances the optimizer, rejected observations contribute a zero current gradient while momentum still decays. The mean force is then the equivalent-risk gradient without the factor $1/r$. These are different conventions for evolving optimizer state, rather than two descriptions of an identical discrete algorithm.

\section{Summary}
The equivalent loss makes the objective induced by uncertainty sampling explicit. Probabilistic, margin-based, and threshold-based acquisition rules can then be studied through the same sequence of questions: calibration identifies their statistical target, curvature determines whether optimization is tractable, and a surrogate link translates optimization error into classification error. The concrete examples show how selective querying can preserve a calibrated convex loss or introduce flat regions and non-convexity even when the original loss is convex.

With a fixed step size $cD/(G\sqrt T)$, the expected query rate enters the stochastic-gradient variance directly. The resulting bound exhibits a transparent tradeoff between initial error and queried-gradient noise. Choosing $c=U_{\text{max}}^{-1/2}$ requires only the largest query probability and produces a leading classification upper bound no larger than the passive-learning counterpart at the same expected label budget. Calibrating $c$ to the average rate can sharpen that comparison. These results give a concrete role to uncertainty beyond selecting labels: its magnitude informs the learning-rate scale, while its variation across observations determines the objective and the possible label-efficiency gain.

The extensions identify where the same perspective must be supplemented. Pool sampling introduces an adaptive normalizer and empirical-to-population error; vector-valued prediction requires integrability of the weighted field; and momentum makes the successful-query clock part of the dynamics. Together, the results provide a reusable framework for analyzing existing acquisition rules and for designing their optimization procedures.

\FloatBarrier
\bibliographystyle{informs2014}
\bibliography{main}

\clearpage
\appendix

\section{Additional Pool-Based Details}\label{apd:pool-details} 

For the pool-based setting in our paper, we can also consider a repeated-query setting where the learner may query the same sample $X_i$ multiple times, and upon each query, the label is generated independently following $\mathcal{P}_{Y|X}$. Practically, this simulates a situation where different human experts may provide different labels for the same sample feature $X$. We can marginalize out the randomness of $Y$ and define the objective as
\[
L(\theta; X) \coloneqq \mathbb{E}_{Y}[\ell(\theta; (X, Y))|\theta, X], \quad \tilde{L}(\theta; X) \coloneqq \mathbb{E}_{Y}[\tilde{\ell}(\theta; (X, Y))|\theta, X].
\]

What impact does this approach have on our theoretical analysis? We go back to the three terms in the previous subsection: optimization, generalization, and approximation. The repeated-query setting is entailed by the objective \eqref{eqn:pool-based-obj}, which optimizes the empirical conditional loss that marginalizes out $Y$.
\begin{equation}
\label{eqn:pool-based-obj}\mathbb{E}\left[\hat{L}(\theta; X)\right] \coloneqq \frac{1}{n}\sum_{i=1}^n\left(\mathbb{E}_{Y}\left[\ell(\theta; (X, Y))\middle| X=X_i\right]\right)
\end{equation}
Again, we omit the discussion on the approximation term. For the optimization term, the argument is almost the same except for only replacing $\hat{\tilde{\ell}}$ by $\hat{\tilde{L}}$. Note that all three assumptions in Proposition~\ref{prop:SGD_converge_pool} remain the same for $\hat{\tilde{L}}$, so the analysis will induce the same result. For the generalization term, the repeated query allows us to marginalize the noises in the sampling procedure of $\mathcal{P}_{Y|X}$. Imagine an extreme case where the query can be made indefinitely many for one single $X_i$. Then, the single query setting cannot capture $\mathcal{P}_{Y|X}$, while the repeated query's distribution will converge to it. In fact, in more general cases, in repeated query setting, we can get a tighter Rademacher complexity and hence a tighter generalization result compared to single query. To be more specific, let's consider for each $X_i$, the repeated-query setting has independently queried $X_i$'s label for $m$ times, resulting in $Y_{i, 1}^{\text{rep}}, \dots, Y_{i, m}^{\text{rep}}$, and the single-query setting has queried $X_i$'s label also for $m$ times but with $Y_{i,1}^{\text{sin}} = \cdots = Y_{i, m}^{\text{sin}}$. We omit the superscript sometimes when there is no ambiguity. Now, we can define the loss class for each setting as follows:
\[
\mathcal{L}^{\text{rep}} \coloneqq \{(x, y_1, \dots, y_m) \mapsto \frac1m \sum_{j=1}^m \ell(f_\theta(x), y_j)| f_\theta \in \mathcal{F}\},
\]
\[
\mathcal{L}^{\text{sin}} \coloneqq \{(x, y) \mapsto \ell(f_\theta(x), y)| f_\theta \in \mathcal{F}\}.
\]
Let's take the expectation of the empirical Rademacher complexity of the repeated-query loss class:
\[
\mathbb{E}_{X_i, Y_{i,j}}\big[\mathcal{R}_n(\mathcal{L}^{\text{rep}})\big] = \mathbb{E}_{X_i, Y_{i,j}, \sigma_i}\bigg[\sup_{f_\theta \in \mathcal{F}} \frac1n \sum_{i=1}^n \sigma_i \frac1m\sum_{j=1}^m \ell(f_\theta(X_i), Y_{i,j})\bigg].
\]
Introduce independent random indices $J_1, \dots, J_m$ each uniform on $\{1, \dots, m\}$, independent of everything else. Then
\[
\frac1m\sum_{j=1}^m \ell(f_\theta(X_i), Y_{i,j}) = \mathbb{E}_{J_i}[\ell(f_\theta(X_i), Y_{i, J_i}) | X_i, Y_{i,1}, \dots, Y_{i,m}].
\]
Hence, we can rewrite the Rademacher complexity as
\[
\mathbb{E}_{X_i, Y_{i,j}}\big[\mathcal{R}_n(\mathcal{L}^{\text{rep}})\big] = \mathbb{E}_{X_i, Y_{i,j}, \sigma_i}\bigg[\sup_{f_\theta \in \mathcal{F}} \frac1n \sum_{i=1}^n \sigma_i \mathbb{E}_{J_i}\big[\ell(f_\theta(X_i), Y_{i, J_i}) \big| X_i, Y_{i,1}, \dots, Y_{i,m}\big]\bigg].
\]
Since for any $H$, $\sup_{f_\theta} \mathbb{E}_J H(f_\theta, J) \leq \mathbb{E}_J \sup_{f_\theta} H(f_\theta, J)$, we have
\[
\mathbb{E}_{X_i, Y_{i,j}}\big[\mathcal{R}_n(\mathcal{L}^{\text{rep}})\big] \leq \mathbb{E}_{X_i, Y_{i,j}, \sigma_i, J_i}\bigg[\sup_{f_\theta \in \mathcal{F}} \frac1n \sum_{i=1}^n \sigma_i \ell(f_\theta(X_i), Y_{i, J_i}) \bigg].
\]
However, $Y_{i,J_i} \sim \mathcal{P}_{Y|X_i}$, which follows the same rule as the single-query setting. Therefore,
\[
\mathbb{E}_{X_i, Y_{i,j}}\big[\mathcal{R}_n(\mathcal{L}^{\text{rep}})\big] \leq \mathbb{E}_{X_i, Y_{i,j}}\big[\mathcal{R}_n(\mathcal{L}^{\text{sin}})\big].
\]

In summary, following the standard Rademacher complexity arguments, we can see the repeated-query setting brings additional benefit to the generalization term in the framework.

\section{Additional Surrogate and Convexity Details}
\subsection{The surrogate-link theorem}\label{apd:bartlett}
\begin{theorem}[Theorem 1 in \citet{bartlett2006convexity}]
\label{thm:bartlett}
For any loss function $\ell$ that can be expressed as $\ell(\hat{Y}, Y)$, one can construct a link function $\psi$. Furthermore, the constructed link function is mini-max optimal in the sense that for any non-negative loss $\ell$, any $|\mathcal{X}| \geq 2$, any risk level $\zeta \in [0, 1]$, and any precision $\varepsilon > 0$, there exists a probability distribution on $\mathcal{X} \times \{-1, +1\}$ such that $L_{\mathrm{01}}(f) - \inf_{g \in \mathcal{G}} L_{\mathrm{01}}(g) = \zeta$ and
\[\psi(\zeta) \leq \mathbb{E}\left[\ell(f(X), Y) \right] -\inf_{g \in \mathcal{G}} \mathbb{E}\left[\ell(g(X), Y)\right] \leq \psi(\zeta)+\varepsilon.\]
The loss $\ell$ is classification-calibrated (Fisher consistent) if and only if for any $z \in (0, 1]$, $\psi(z) > 0.$
\end{theorem}
\subsection{Additional convexity calculations}\label{apd:convex-details}
The formulas below give additional details for the sample-level equivalent losses used in the main text.
For the exponential and ordinary-margin constructions, see the explicit antiderivatives in Appendix~\ref{apd:equiv_loss} and the curvature calculations below. For logistic parameterization, the least-confidence derivative at $z<0$, $Y=1$ is $-q(1-q)$, with $q=(1+e^{-z})^{-1}$; its derivative is $-q(1-q)(1-2q)$, which is negative.

\section{Derivation of Equivalent Losses and Surrogate Link Function }
This section will present the detailed calculations of the equivalent losses and the surrogate link functions of all the listed examples in previous sections. The subscript of $\mu$ or $\gamma$ will sometimes be omitted for simplicity when the text is clear.
\subsection{Equivalent Loss in Section \ref{subsec:equiv_loss}}
\label{apd:equiv_loss}
\setcounter{example}{0}
\begin{example}[Equivalent loss of probabilistic model]
Both the loss and the uncertainty function can be expressed as a function of predicted probability $q(X;\theta)$. By the chain rule,
\[\frac{\partial \tilde{\ell}}{\partial \theta} = \frac{\partial \tilde{\ell}}{\partial q} \cdot \frac{\partial q}{\partial \theta},\]
\[U\cdot \frac{\partial l}{\partial \theta} = U \cdot \frac{\partial l}{\partial q} \cdot \frac{\partial q}{\partial \theta}.\]
Hence if we can find some $\tilde{\ell}$ such that
\[\frac{\partial \tilde{\ell}}{\partial q} = U \cdot \frac{\partial l}{\partial q},\]
then we have accomplished the task.\\
The indicator function $\mathbbm{1}\{Y=+1\}$ where $Y\in \{-1, +1\}$ can be transformed into $\frac{Y+1}{2}$ which we denote as $p$ by a slight abuse of notation. Then the derivative of the original cross-entropy loss can be presented as
\[\frac{\partial l}{\partial q} = -\frac{p}{q} + \frac{1-p}{1-q} = \frac{q-p}{q(1-q)}.\]
We start with the entropy uncertainty $U = -q \log(q) - (1-q) \log(1-q)$ in \citet{dagan1995committee}.
\allowdisplaybreaks
\begin{align*}
U(q) \cdot \frac{\partial l}{\partial q} & = p \log(q) -(1-p) \log(1-q) - (1-p)\cdot\frac{q \log(q)}{1-q} + p \cdot \frac{(1-q) \log(1-q)}{q} \\
& = p \log(q) -(1-p) \log(1-q) \\
& \phantom{=} - (1-p)\cdot\frac{(q - 1) \log(q) + \log(q)}{1-q} + p \cdot \frac{-q \log(1-q) + \log(1-q)}{q} \\
& = p \log(q) -(1-p) \log(1-q) + (1-p) \log(q) \\
& \phantom{=}- (1-p)\cdot\frac{\log(q)}{1-q} - p \log(1-q) + p \cdot \frac{\log(1-q)}{q}\\
& = \log(q) - \log(1-q) - (1-p)\cdot\frac{\log(q)}{1-q} + p \cdot \frac{\log(1-q)}{q}.
\end{align*}
Then by calculating its indefinite integral, we have
\[\int U(q) \cdot \frac{\partial l}{\partial q} \mathrm{d}q = q\log(q) + (1-q)\log(1-q) - p \cdot \mathrm{Li}_2(q) - (1-p) \cdot \mathrm{Li}_2(1-q) + C,\]
where $\mathrm{Li}_2(z)$ is the Spence's function,
\[\mathrm{Li}_2(z) = -\int_{0}^{z} \frac{\log(1-z)}{z} \mathrm{d}z.\]
Since we are interested in the excess risk (which is the expected difference between those hypotheses and the optimal measurable function), the selection of $C$ does not matter. We simply select $C = \mathrm{Li}_2(1) = \frac{\pi^2}{6}$ to make the equivalent loss vanish at $p = q = 0$ and $p = q = 1$, which yields the equivalent loss $\tilde{\ell}$ presented in Section \ref{subsec:equiv_loss}.

We continue with the least confident uncertainty $U = \min\{q, 1-q\}$ in \citet{culotta2005reducing}. For $q \in [0, \frac{1}{2}]$, we have
\allowdisplaybreaks
\begin{align*}
U(q) \cdot \frac{\partial l}{\partial q} & = \frac{q-p}{1-q} \\
& = \frac{q - 1 + 1 - p}{1-q}\\
& = -1 + \frac{1-p}{1-q}.
\end{align*}
Its indefinite integral is simple:
\[\int U(q)\cdot \frac{\partial l}{\partial q} \mathrm{d}q = -q - (1-p) \log(1-q) + C, \quad \forall q \in [0, 0.5].\]
Similarly, we can compute the indefinite integral for $q \in [0.5, 1]$:
\[\int U(q)\cdot \frac{\partial l}{\partial q} \mathrm{d}q = q - p \log(q) + C, \quad \forall q \in [0.5, 1].\]
The equivalent loss function is piecewise continuous. We select the constants properly to avoid the jump discontinuity at $q = \frac12$.
To let the values at $q = \frac12$ match each other, we select the constants so that
\[\tilde{\ell} = \begin{cases}
-(1-p)\cdot\log(2(1-q)) - q + \log(2), \quad & \text{if } q < 0.5;\\
-p\cdot\log(2q) - (1-q) +\log(2), \quad & \text{if } q \geq 0.5.
\end{cases}\]
Again, we don't choose the overall constant deliberately. The $\log(2)$ term is simply to make the equivalent loss vanish at $p = q = 0$ and $p = q = 1$.
\end{example}

\begin{example}[Equivalent loss of margin-based model]
For the SVM-based methods, both the loss and the uncertainty function can be expressed as a function of $Y \cdot \hat{Y}$, where $\hat{Y} =  \theta^\top X$. By the similar chain rule arguments in Example \ref{eg:probabilistic}, we can find the equivalent loss with respect to $\theta$ as long as we can find that with respect to $Y \cdot \hat{Y}$. 
To simplify the notations, we denote $Y\cdot \hat{Y} = Y \theta^\top X$ by $s$. As a reminder, we again state the squared Hinge loss
\[\ell(s) = \begin{cases}
(1-s)^2, &\quad \text{if } s \leq 1;\\
0, &\quad \text{if } s \geq 1,
\end{cases}\]
and the uncertainty function
\[U_{\mu}(s) = \begin{cases}
(1-\mu s)^{-1}, &\quad \text{if } s \leq 0;\\
(1+\mu s)^{-1}, &\quad \text{if } s \geq 0.
\end{cases}\]
We compute the amount $U \cdot \frac{\partial l}{\partial s}$ and its indefinite integral in three parts.\\
For $s \geq 1$, the result is straightforward: the equivalent loss must be a constant. We select the constant to be zero for some notation convenience.\\
For $s \in [0, 1]$,
\allowdisplaybreaks
\begin{align*}
U_{\mu}(s) \cdot \frac{\partial l}{\partial s} & = -2(1-s) \cdot \frac{1}{1+\mu s}\\
& = -2 \frac{-\frac1\mu (\mu s + 1) + \frac1\mu + 1}{1+\mu s}\\
& = \frac2\mu - 2(\frac1\mu + 1) \cdot \frac{1}{1+\mu s}.
\end{align*}
Its indefinite integral is
\[\int U_{\mu}(s) \cdot \frac{\partial l}{\partial s} \mathrm{d} s = \frac2\mu \cdot s - \frac2\mu (\frac1\mu + 1) \cdot \log(1+\mu s) + C,\]
where we select $C = \frac2\mu- \frac2\mu (\frac1\mu + 1) \cdot \log(1+\mu)$ so that the values at $s = 1$ coincide.\\
For $s \leq 0$, we can complete the calculation similarly:
\allowdisplaybreaks
\begin{align*}
U_{\mu}(s) \cdot \frac{\partial l}{\partial s} & = -2(1-s) \cdot \frac{1}{1-\mu s}\\
& = -2 \frac{-\frac1\mu (1 - \mu s) - \frac1\mu + 1}{1-\mu s}\\
& = -\frac2\mu + 2(\frac1\mu - 1) \cdot \frac{1}{1-\mu s}.
\end{align*}
The indefinite integral is
\[\int U_{\mu}(s) \cdot \frac{\partial l}{\partial s} \mathrm{d} s = -\frac2\mu \cdot s - \frac2\mu (\frac1\mu - 1) \cdot \log(1-\mu s) + C,\]
where the constant is selected to be the same as $s \in [0, 1]$ to match at $s = 0$.
\end{example}

\begin{example}[Equivalent loss of threshold-based model]
The uncertainty function is probably the simplest case: an indicator function of whether $|s| = |Y\cdot\hat{Y}| = |Y \theta^\top X|$ is no greater than a certain threshold $\gamma$. Then for those $s$'s that satisfy the threshold requirement, the equivalent loss is identical to the original loss (which is the logistic loss, as a reminder), while for those $s$'s outside the threshold area, the equivalent loss must be constant. We select those constants to avoid abrupt changes at the threshold, resulting in the expressions in Section \ref{subsec:equiv_loss}.
\end{example}

\begin{example}[Equivalent loss of margin loss and margin-based uncertainty]
We recall that the original loss and the uncertainty function with respect to $s = \hat{Y}\cdot Y$ are
\[\ell(s) = \max\{0, 1-s\},\]
\[U_{\mu}(s) = \begin{cases}
\frac{1}{1+\mu s}, & \quad \text{if } s \geq 0;\\
\frac{1}{1-\mu s}, & \quad \text{if } s \leq 0.
\end{cases}\]
For the $s \geq 1$ part, the indefinite integral must be constant. We select the constant to be zero.\\
For the $s \in (0, 1)$ part,
\[U_{\mu}(s) \cdot \frac{\partial l}{\partial s} = -\frac{1}{1+\mu s},\]
which indicates that
\[\int U_{\mu}(s) \cdot \frac{\partial l}{\partial s} \mathrm{d}s = -\frac{1}{\mu} \log(1+\mu s) + C, \quad \forall s \geq 0. \]
We select the constant to be $\frac{1}{\mu} \log(1+\mu)$ so that there is no discontinuity at $s = 1$.\\
For the $s \leq 0$ part,
\[U_{\mu}(s) \cdot \frac{\partial l}{\partial s} = -\frac{1}{1-\mu s},\]
resulting in
\[\int U_{\mu}(s) \cdot \frac{\partial l}{\partial s} \mathrm{d}s = \frac{1}{\mu} \log(1-\mu s) + C, \quad \forall s \geq 0. \]
We set the constant to be $\frac{1}{\mu} \log(1+\mu)$ to keep the continuity at $s = 0$.
\end{example}

\begin{example}[Equivalent loss of exponential loss and exponential uncertainty]
Similarly, we state the original loss and the uncertainty function concerning $s$:
\[\ell(s) = \exp(-s),\]
\[U_{\mu}(s) = \begin{cases}
\exp(-\mu s), & \quad \text{if } s \geq 0;\\
\exp(\mu s), & \quad \text{if } s \leq 0.
\end{cases}\]
Then, for $s \geq 0$,
\[U_{\mu}(s) \cdot \frac{\partial l}{\partial s} = -\exp(-(1+\mu) s),\]
of which the indefinite integral is
\[\int U_{\mu}(s) \cdot \frac{\partial l}{\partial s} \mathrm{d} s = \frac{\exp(-(1+\mu)s)}{1+\mu} + C, \quad \forall s \geq 0.\]
We select $C = \frac{\mu}{1+\mu}$ so that the value at $s = 0$ is 1.\\
On the contrary, for $s \leq 0$,
\[U_{\mu}(s) \cdot \frac{\partial l}{\partial s} = -\exp(-(1-\mu) s),\]
leading to
\[\int U_{\mu}(s) \cdot \frac{\partial l}{\partial s} \mathrm{d} s = \frac{\exp(-(1-\mu)s)}{1-\mu} + C, \quad \forall s \leq 0.\]
The constant is chosen to be $C = -\frac{\mu}{1-\mu}$ to meet the value at $s=0$.
\end{example}

\subsection{Surrogate Property and Proof of Proposition \ref{prop:link_calculate}}
\label{apd:surrogate}
In this subsection, we summarize the arguments in \citet{bartlett2006convexity} and provide their surrogate link function computation method for the margin-based models such as the SVM. Such a surrogate property induces a minimax optimal bound on the excessive 0-1 risk (see Theorem 3 in \citet{bartlett2006convexity}). For simplicity, in this subsection, we omit the dependence on $X$ and $\theta$, since all the excess risk analyses hold for any certain but fixed hypothesis $f_\theta$ and sample point $X=x$.

We start with the standard definitions of \citet{bartlett2006convexity}. Assume that the loss $\ell(\hat{Y}, Y)$ is of the form $\ell(\hat{Y} \cdot Y)$ (which is the case in all of our examples). By denoting the probability of a positive $Y$ by $p$, the expected loss induced by predicting $\hat{Y}$ is
\[C_p(\hat{Y}) \coloneqq p \ell(\hat{Y}) + (1-p) \ell(-\hat{Y}).\]
For any fixed probability value $p$, the inferior of the expected loss is denoted by
\[H(p) \coloneqq \inf_{\hat{Y}} \, C_p(\hat{Y}).\]
If we restrict the prediction $\hat{Y}$ to be not Bayes-optimal (that is, to be of the different sign as $2p-1$) and take the inferior, we get
\[H^-(p) \coloneqq \inf_{\hat{Y} \cdot (2p - 1)\leq 0}\, C_p(\hat{Y}).\]
Note that a binary classification loss $\ell$ is said to be classification-calibrated \citep{bartlett2006convexity} (or Fisher consistent \citep{lin2004note}) if $H^-(p) > H(p)$ for any $p \neq \frac{1}{2}$.

\citet{bartlett2006convexity} provide a way of computing the surrogate link function $\psi: [0, 1] \rightarrow \mathbb{R}$ via
\[\bar{\psi}(z) = H^-\left(\frac{1+z}{2}\right) - H\left(\frac{1+z}{2}\right),\]
\[\psi(z) = \bar{\psi}^{**}(z),\]
where $g^{**}$ is the Fenchel-Legendre biconjugate of the function $g$, characterized by
\[\mathrm{epi} \, g^{**} = \overline{\mathrm{co}} \phantom{=} \mathrm{epi}\, g.\]
Note that those functions are convex if and only if their Fenchel-Legendre biconjugate are themselves \citep{bartlett2006convexity}.

Equipped with such a surrogate link function $\psi$, \citet{bartlett2006convexity}'s Theorem 1 
shows that it can be an upper bound for the excessive 0-1 risk: for any measurable function $f$ and any probability distribution on $\mathcal{X} \times \mathcal{Y} = \mathcal{X} \times \{-1, +1\}$,
\[\psi\left(L_{\mathrm{01}}(f) - \inf_{g \in \mathcal{G}} L_{\mathrm{01}}(g) \right) \leq \mathbb{E}\left[\ell(f(X), Y) \right] -\inf_{g \in \mathcal{G}} \mathbb{E}\left[\ell(g(X), Y)\right],\]
where $\mathcal{G}$ is the set of all measurable functions.

Such an upper bound is minimax optimal in the sense that for any non-negative loss $\ell$, any $|\mathcal{X}| \geq 2$, any 0-1 risk level $\zeta \in [0, 1]$, and any precision $\varepsilon > 0$, there exists a probability distribution on $\mathcal{X} \times \{-1, +1\}$ such that $L_{\mathrm{01}}(f) - \inf_{g \in \mathcal{G}} L_{\mathrm{01}}(g) = \zeta$, and
\[\psi(\zeta) \leq \mathbb{E}\left[\ell(f(X), Y) \right] -\inf_{g \in \mathcal{G}} \mathbb{E}\left[\ell(g(X), Y)\right] \leq \psi(\zeta)+\varepsilon.\]

Equipped with such powerful tools, all we need to do is to find the surrogate link functions of those active learning models. But before we proceed to the particular calculation, we notice that the analysis in \citet{bartlett2006convexity} is designed for the margin-based models, while our Example \ref{eg:probabilistic} is not based on the margin but on the probability. To generalize the arguments to the probabilistic models, we transform the probability into the expectation to enable the margin-based analysis. We denote the predicted expectation of $Y$ in a probabilistic model by 
\[\hat{Y} \coloneqq \hat{\mathbb{E}}[Y] = 2 q - 1.\]
\setcounter{example}{0}
\begin{example}[Surrogate link function of probabilistic model]
Remind that the original loss can be expressed as
\[\ell(\hat{Y} \cdot Y) = -\log\left(\frac{1+\hat{Y} \cdot Y}{2}\right).\]
The entropy uncertainty is
\allowdisplaybreaks
\begin{align*}
U &= -[q\log(q) + (1-q) \log(1-q)]\\
&= -\left[\frac{1+\hat{Y}}{2} \log\left(\frac{1+\hat{Y}}{2}\right) + \frac{1-\hat{Y}}{2} \log\left(\frac{1-\hat{Y}}{2}\right)\right]\\
&= -\left[\frac{1+\hat{Y} \cdot Y}{2} \log\left(\frac{1+\hat{Y} \cdot Y}{2}\right) + \frac{1-\hat{Y} \cdot Y}{2} \log\left(\frac{1-\hat{Y} \cdot Y}{2}\right)\right].
\end{align*}
Then the equivalent loss is
\[\tilde{\ell} = \mathrm{Li}_2(1) - \mathrm{Li}_2\left(\frac{1+ \hat{Y}\cdot Y}{2}\right) + \frac12\left[(1+\hat{Y} \cdot Y) \log\left(\frac{1+\hat{Y} \cdot Y}{2}\right) + (1-\hat{Y} \cdot Y) \log\left(\frac{1-\hat{Y} \cdot Y}{2}\right)\right],\]
where $\mathrm{Li}_2(\cdot)$ is the Spence's function.
One can take an easy check that this loss is actually identical to the equivalent loss we provide in Section \ref{subsec:equiv_loss} if $\hat{Y} = 2q-1$.

Notice that $U$ is a non-negative even function that only takes zero value at two endpoints, which implies that minimizing expected $\tilde{\ell}$ is equivalent to minimizing expected $\ell$. The minimizer $\hat{Y}^*$ can be easily obtained at the first-order stationary point
\[p \cdot \left(-\frac{1}{1+\hat{Y}^*}\right) + (1-p) \cdot \left(\frac{1}{1-\hat{Y}^*}\right) = 0,\]
which is $\hat{Y}^* = 2p - 1$. Then
\[H(p) = \mathrm{Li}_2(1) -\left[p \mathrm{Li}_2(p) + (1-p) \mathrm{Li}_2(1-p)\right] + \left[p\log(p) + (1-p)\log(1-p)\right],\]
where $\mathrm{Li}_2(\cdot)$ is the Spence's function.

The computation of $H^-(p)$ is simple: the equivalent loss is convex, indicating that the inferior risk of the non-Bayes classifiers must be taken at $\hat{Y} = 0$. Therefore,
\[H^-(p) = \mathrm{Li}_2(1) -\mathrm{Li}_2\left(\frac12\right) - \log(2).\]

By definition,
\begin{align*}
\bar{\psi}(z) & = -\mathrm{Li}_2\left(\frac12\right) + \frac{1+z}{2}\cdot \mathrm{Li}_2\left(\frac{1+z}{2}\right) + \frac{1-z}{2}\cdot \mathrm{Li}_2\left(\frac{1-z}{2}\right) \\
& \phantom{=}- \left[\frac{1+z}{2}\cdot \log(1+z) + \frac{1-z}{2}\cdot \log(1-z)\right],
\end{align*}
whose second-order derivative is
\[\frac{\mathrm{d}^2 \bar{\psi}}{\mathrm{d} z^2} = -\frac12\left[\frac{(1-z)\log\left(\frac{1-z}{2}\right) + (1+z) \log\left(\frac{1+z}{2}\right) }{1-z^2}\right] \geq 0.\]
The convexity implies that
\[\psi(z) = \bar{\psi}(z).\]

We need to note that the first-order derivative of $\psi$ is
\[\frac{\mathrm{d}\psi}{\mathrm{d}z} = \frac12 \left[\mathrm{Li}_2\left(\frac{1+z}{2}\right) - \mathrm{Li}_2\left(\frac{1-z}{2}\right)\right] \geq 0,\]
which is zero if and only if $z=0$. So the equivalent loss is classification-calibrated, and the surrogate link function around $z=0$ is approximately
\[\psi(z) \sim \frac12\frac{\mathrm{d}^2 \psi}{\mathrm{d} z^2}\Bigg\vert_{z=0} \cdot z^2 = \frac{\log(2)}{2} \cdot z^2.\]
Since $\psi(z)$ is bounded at $z\in[0,1]$, we can conclude that $\psi(z) = \Theta(z^2),$
where $\Theta$ is the big theta notation referring to ``of the same order as'' rather than our denoted set of parameters.

The other example of the least confidence uncertainty $U = \min\{q, 1-q\}$ can also be analyzed via $\hat{Y} = 2q -1$. By definition,
\[U = \min\left\{\frac{1+\hat{Y}}{2}, \frac{1-\hat{Y}}{2}\right\} = \frac{1-|\hat{Y}|}{2}.\]
The equivalent loss with respect to $\hat{Y} \cdot Y$ is
\[\tilde{\ell}(\hat{Y} \cdot Y) = \begin{cases}
\frac12 \big(\hat{Y}\cdot Y -2\log(1+\hat{Y}\cdot Y)\big) + \log(2) - \frac12, &\quad \text{if } \hat{Y}\cdot Y \geq 0;\\
-\frac12 \cdot \hat{Y}\cdot Y + \log(2) - \frac12, & \quad \text{if } \hat{Y}\cdot Y \leq 0.
\end{cases}\]
Again, one can quickly check that this equivalent loss is identical to the form we present in Section \ref{subsec:equiv_loss} with $\hat{Y} = 2q - 1$. We don't bother to adjust those constants explicitly to meet the non-negativity or any other requirements, since those equivalent losses are all bounded and we are interested in the excess risk (which is one expected loss minus another).

W.l.o.g. assume that $p \geq \frac12$. Then the first-order stationary point of $C_p(\hat{Y})$ should be
\[-\frac12 \cdot p \cdot \frac{1-\hat{Y}^*}{1+\hat{Y}^*} + \frac12 (1-p) = 0,\]
which is $\hat{Y}^* = 2p - 1$. Then
\begin{align*}
H(p) &= p \cdot \frac12\left[ (2p-1) - 2\log(2 p) \right] - (1-p) \cdot \frac12(1-2p) + \log(2) - \frac12 \\
&= p - \frac12 - p \log(2p) + \log(2) - \frac12, \quad \forall p \geq \frac12.
\end{align*}
For $p \leq \frac12$, the optimal $\hat{Y}^*$ remains the same $2p - 1$, while $\forall p \leq \frac12$,
\begin{align*}
H(p) & = -p \cdot \frac12(2p - 1) + (1-p) \cdot \frac12\left[ (1-2p) - 2 \log(2(1-p))\right] + \log(2) - \frac12 \\
& = \frac12 - p - (1-p) \log(2(1-p)) + \log(2) - \frac12.
\end{align*}
By the convexity of $\tilde{\ell}$,
\[H^-(p) = C_p(0) = \log(2) - \frac12.\]
The derivation of $\bar{\psi}$ only requires the $p \geq \frac12$ part, hence
\[\bar{\psi}(z) = H^-\left(\frac{1+z}{2}\right) - H\left(\frac{1+z}{2}\right) = -\frac12 z + \frac{1+z}{2}\log(1+z),\]
of which the second-order derivative is
\[\frac{\mathrm{d}^2 \bar{\psi}}{\mathrm{d} z^2} = \frac{1}{2(1+z)} \geq 0.\]
By the convexity of $\bar{\psi}$, we have
\[\psi= \bar{\psi}.\]

From the first-order derivative of $\psi$
\[\frac{\mathrm{d} \psi}{\mathrm{d}z} = \frac12 \log(1+z),\]
we know that the surrogate link function $\psi$ is only tending to zero if and only if $z$ itself tends zero. Thus, the equivalent loss is classification-calibrated. From the facts that
\[\frac{\mathrm{d} \psi}{\mathrm{d}z}\Bigg\vert_{z=0} = 0\]
and
\[\frac{\mathrm{d}^2 \psi}{\mathrm{d}z^2}\Bigg\vert_{z=0} = \frac12,\]
we know that
\[\psi(z) \sim \frac14 z^2\]
around the zero point. From the boundedness of $\psi$, we can also conclude similarly to the entropy uncertainty case that
\[\psi(z) = \Theta(z^2),\]
where the big theta notation means ``of the same order as''.
\end{example}

\begin{example}[Surrogate link function of margin-based model]
We start with finding the $\hat{Y}$ that minimizes the expected equivalent loss. Remind that the equivalent loss can be written in the form of $\hat{Y} \cdot Y$:
\[\tilde{\ell}_{\mu} = \begin{cases}
-\frac{2}{\mu}(\frac{1}{\mu} - 1)\log(1-\mu \hat{Y} \cdot Y) - \frac{2}{\mu} \hat{Y} \cdot Y + C,\quad & \text{if } \hat{Y} \cdot Y \leq 0;\\
-\frac{2}{\mu}(\frac{1}{\mu} + 1)\log(1+\mu \hat{Y} \cdot Y) + \frac{2}{\mu} \hat{Y} \cdot Y + C,\quad & \text{if } \hat{Y} \cdot Y \in (0, 1);\\
0, \quad & \text{if } \hat{Y} \cdot Y \geq 1,
\end{cases}\]
where $C = \frac{2}{\mu}(\frac{1}{\mu} + 1)\log(1+\mu) - \frac{2}{\mu}$.
By the definition,
\[U_\mu \cdot \frac{\partial l}{\partial \hat{Y}} = \frac{\partial \tilde{\ell}_\mu}{\partial \hat{Y}}.\]
Since $U_\mu = (1+\mu |\hat{Y}|)^{-1}$ is a positive and even function, minimizing the expected equivalent loss is identical to minimizing the expected original loss (which is, the squared Hinge loss). By direct calculation (or referring the Example 2 in \citet{bartlett2006convexity}), the minimizer should be
\[\hat{Y}^* = 2p - 1.\]
Without loss of generality, we assume $p \geq \frac12$, which implies that $2p-1 \geq 0$. Subject to that minimizer,
\begin{align*}
H(p) &= C \\
&\, + p \cdot \left(-\frac2\mu \left(\frac1\mu + 1\right) \log(1+\mu (2p - 1)) + \frac2\mu (2p-1)\right) \\
&\, + (1-p)\cdot \left(-\frac2\mu \left(\frac1\mu -1\right) \log(1+\mu(2p-1))+\frac2\mu (2p-1)\right). 
\end{align*}
Since the equivalent loss is convex, the minimized risk of the non-Bayes classifier must be
\[H^-(p) = C_p(0) = C.\]
Hence we have
\begin{align*}
\bar{\psi}(z) & = H^-\left(\frac{1+z}{2}\right) - H\left(\frac{1+z}{2}\right)\\
& = \frac{1+z}{2}\cdot \left(\frac2\mu \left(\frac1\mu + 1\right) \log(1+\mu z) - \frac2\mu z\right) + \frac{1-z}{2}\cdot \left(\frac2\mu \left(\frac1\mu -1\right) \log(1+\mu z)-\frac2\mu z\right)\\
& = \frac{2}{\mu^2} \cdot (1+\mu z) \log(1+\mu z) - \frac2\mu z.
\end{align*}
The second-order derivative of $\tilde(\psi)$ is
\[\frac{\mathrm{d}^2 \bar{\psi}}{\mathrm{d} z^2} = \frac{2}{1+\mu z} > 0,\]
which guarantees the convexity of $\bar{\psi}$. Hence
\[\psi = \bar{\psi}.\]
The first-order derivative of $\psi$ is
\[\frac{\mathrm{d} \psi}{\mathrm{d} z} = \frac2\mu \log(1+\mu z) \geq 0,\]
where the equality holds if and only if $z = 0$ for any $\mu > 0$, indicating the classification-calibration of the equivalent loss $\tilde{\ell}$. By a similar Taylor expansion argument, we can conclude that
\[\psi(z) \sim \frac12 \cdot \frac{\mathrm{d}^2 \psi}{\mathrm{d} z^2} \Bigg\vert_{z=0} \cdot z^2 = z^2.\]
Due to the boundedness of the surrogate link function, we have
\[\psi(z) = \Theta(z^2),\]
where the big theta notation stands for ``of the same order as''.
\end{example}

\begin{example}[Surrogate link function of threshold-based model]
We briefly recall the equivalent loss with respect to $\hat{Y}$
\[\tilde{\ell}_{\gamma} = \begin{cases}
\log(1+\exp(\gamma)),\quad & \text{if } Y \cdot \hat{Y} \leq -\gamma;\\
\log(1+\exp(-Y \cdot \hat{Y})),\quad & \text{if } Y \cdot \hat{Y} \in (-\gamma, \gamma);\\
\log(1+\exp(-\gamma)), \quad & \text{if } Y \cdot \hat{Y} \geq \gamma,
\end{cases}\]
where the non-constant part is identical to that of a logistic loss. For sufficiently large threshold $\gamma$ so that the minimizer locates in the non-constant part, we compute the first-order condition of the minimizer (which is just that of the logistic loss) as
\[-p \cdot \frac{\exp(-\hat{Y}^*)}{1+\exp(-\hat{Y}^*)} + (1-p) \cdot \frac{\exp(\hat{Y}^*)}{1+\exp(\hat{Y}^*)} = 0,\]
which implies that
\[\hat{Y}^* = \log\left(\frac{p}{1-p}\right).\]
For a small $\gamma$, the derivative of the expected equivalent loss suggests that the minimizer should be
\[\hat{Y}^* = \gamma \cdot \mathrm{sign}(2p - 1).\]
Without loss of generality, we assume that $p \geq \frac12$. Then 
\[\hat{Y}^* = \begin{cases}
\gamma, &\quad \text{if } p \geq \frac{\exp(\gamma)}{1+\exp(\gamma)};\\
\log\left(\frac{p}{1-p}\right), &\quad \text{if } \frac12 \leq p \leq \frac{\exp(\gamma)}{1+\exp(\gamma)}.
\end{cases}\]
Substituting above results into $C_p(\hat{Y})$, we have
\[H(p) = \begin{cases}
-[p\log(p) + (1-p)\log(1-p)], & \quad \text{if } \frac12 \leq p \leq \frac{\exp(\gamma)}{1+\exp(\gamma)};\\
\log(1+\exp(\gamma)) - \frac{\gamma \exp(\gamma)}{1+\exp(\gamma)},&\quad \text{if } p \geq \frac{\exp(\gamma)}{1+\exp(\gamma)}.
\end{cases}\]
One can check that $C_p(0) \leq C_p(\hat{Y})$ for any $p \geq \frac12$ and $\hat{Y} \leq 0$, implying that
\[H^-(p) = C_p(0) = \log(2).\]
Then
\[\bar{\psi}(z) = \begin{cases}
\frac12\left[(1+z) \log(1+z) + (1-z)\log(1-z)\right], &\quad \text{if } z \leq \frac{\exp(\gamma) - 1}{\exp(\gamma) + 1};\\
\log\left(\frac{2}{1+\exp(\gamma)}\right) + \frac{\gamma \exp(\gamma)}{1+\exp(\gamma)},  &\quad \text{if } z \geq \frac{\exp(\gamma) - 1}{\exp(\gamma) + 1}.
\end{cases}\]
Apparently, $\bar{\psi}(z)$ is non-convex as a whole: in the first part where $z$ is small, the function is convex and strictly increasing, while in the second part, the function is a constant. We extend the values of $\bar{\psi}(z)$ from small $z$'s to large $z$'s by defining another function
\[h(z) \coloneqq \frac12\left[(1+z) \log(1+z) + (1-z)\log(1-z)\right].\]
To compute $\psi(z)$, observe that the convex hull of the epigraph of $\bar{\psi}$ can be determined by some specific point $z_0 \leq \frac{\exp(\gamma) - 1}{\exp(\gamma) + 1}$: at the left side of $z_0$, the epigraph is identical to that of $h(z)$, while at the right side of $z_0$, the epigraph is identical to that of the tangent at $(z_0, h(z_0))$. Such a tangent should contain the right-most point $(1, \bar{\psi}(1))$, which means
\[h(z_0) + h^\prime(z_0)\cdot (1-z_0) = \bar{\psi}(1).\]
Replacing the equation with concrete expressions, we have
\begin{align*}
\log(1+z_0) &= h(z_0) + h^\prime(z_0)\cdot (1-z_0) \\
&= \bar{\psi}(1)\\
&= \log(2) - \log\left(\exp\left(\frac{\gamma}{1+\exp(\gamma)}\right) + \exp\left(\frac{-\gamma}{1+\exp(-\gamma)}\right)\right).
\end{align*}
Simplifying notations, we have
\[z_0 = 2\cdot \left(\exp\left(\frac{\gamma}{1+\exp(\gamma)}\right) + \exp\left(\frac{-\gamma}{1+\exp(-\gamma)}\right)\right)^{-1} - 1. \]
Therefore,
\[\psi(z) =
\begin{cases}
\frac12\left[(1+z) \log(1+z) + (1-z)\log(1-z)\right], & \quad \text{if } z \leq z_0;\\
\frac12\left[(1+z) \log(1+z_0) + (1-z)\log(1-z_0)\right], & \quad \text{if } z \geq z_0,
\end{cases}\]
where $z_0$ is some positive constant stated above.
By examining the first-order derivative of $\psi(z)$, we can easily find out that the equivalent loss is classification-calibrated:
\[\frac{\mathrm{d}\psi}{\mathrm{d}z} = \begin{cases}
\frac12\left[\log(1+z) - \log(1-z)\right], & \quad \text{if } z \leq z_0;\\
\frac12\left[\log(1+z_0) - \log(1-z_0)\right], & \quad \text{if } z \geq z_0.
\end{cases}\]
By computing its Taylor expansions at $z=0$, we have
\[\psi(z) \sim \frac12 \cdot \frac{\mathrm{d}^2 \psi}{\mathrm{d} z^2} \Bigg\vert_{z=0} \cdot z^2 = \frac12 z^2.\]
Finally, we note that
\[\psi(z) = \Theta(z^2),\]
where the big theta notation suggests ``at the same order as''.
\end{example}

\begin{example}[Surrogate link function of margin loss and margin-based uncertainty]
Similarly, the even and positive uncertainty function $U$ leads to the same minimizer of the expected equivalent loss as the expected original margin loss, while the latter by the arguments in \citet{bartlett2006convexity} is
\[\hat{Y}^* = \mathrm{sign}\left(p - \frac12\right),\]
for $p \neq \frac12$. For $p = \frac12$, any $\hat{Y} \in [-1, 1]$ will lead to the same expected equivalent loss.\\
We compute the $p \geq \frac12$ part, gaining
\[H(p) = p \cdot 0 + (1-p) \cdot \frac{2}{\mu} \log(1+\mu) = (1-p) \cdot \frac{2}{\mu} \log(1+\mu), \quad \forall p \geq \frac12.\]
The other part $p < \frac12$ is
\[H(p) = p \cdot \frac{2}{\mu} \log(1+\mu) + (1-p) \cdot 0 = p \cdot \frac{2}{\mu} \log(1+\mu), \quad \forall p < \frac12.\]
For computing the $H^-(p)$, assume that $p \geq \frac12$. Then any $\hat{Y} \in [-1, 0]$ will be optimal among the non-Bayes classifiers, leading to
\[H^-(p) = \frac{2}{\mu}\log(1+\mu).\]
Hence,
\[\bar{\psi}(z) = H^-\left(\frac{1+z}{2}\right) - H\left(\frac{1+z}{2}\right) = \frac{\log(1+\mu)}{\mu} z.\]
The linear function is of course convex, so
\[\psi = \bar{\psi}.\]
\end{example}

\begin{example}[Surrogate link function of exponential loss and exponential uncertainty]
The equivalent loss concerning $\hat{Y} \cdot Y$ is
\[\tilde{\ell} = \begin{cases}
-\frac{1}{1+\mu} \cdot \exp(-(1+\mu) \hat{Y}\cdot Y) + \frac{\mu}{1+\mu}, & \quad \text{if } \hat{Y} \cdot Y \geq 0;\\
-\frac{1}{1-\mu} \cdot \exp(-(1-\mu) \hat{Y}\cdot Y) - \frac{\mu}{1-\mu}, & \quad \text{if } \hat{Y} \cdot Y \leq 0.
\end{cases}\]
Since the uncertainty function $U = \exp(-|\hat{Y}|)$ is even and positive, the minimizer of the expected equivalent loss $C_p(\hat{Y})$ is identical to that of the expected original loss. That is,
\[-p \exp(-\hat{Y}^*) + (1-p) \exp(\hat{Y}^*) = 0,\]
which implies that
\[\hat{Y}^* = \frac12 \log\left(\frac{p}{1-p}\right).\]
Without loss of generality, assume that $p \geq \frac12$. Then
\begin{align*}
H(p) &= p \left[-\frac{1}{1+\mu} \cdot \exp\left(-\frac{1+\mu}{2} \log\left(\frac{p}{1-p}\right)\right)+ \frac{\mu}{1+\mu}\right] \\
& \phantom{=} + (1-p) \left[-\frac{1}{1-\mu} \cdot \exp\left( \frac{1-\mu}{2} \log\left(\frac{p}{1-p}\right)\right) - \frac{\mu}{1-\mu}\right]\\
&= \frac{2}{1-\mu^2} \cdot p^{\frac{1-\mu}{2}} (1-p)^{\frac{1+\mu}{2}} + \frac{\mu}{1-\mu^2} \cdot (2p-1-\mu).
\end{align*}
Since the equivalent loss is convex with respect to $\hat{Y} \cdot Y$, the minimum of expected equivalent loss when the prediction is non-Bayes is
\[H^-(p) = C_p(0) = 1.\]
Then by definition,
\begin{align*}
\bar{\psi}(z) & = H^-\left(\frac{1+z}{2}\right) - H\left(\frac{1+z}{2}\right)\\
& = \frac{1}{1-\mu^2} \left(1-\mu z - (1-z)^{\frac{1+\mu}{2}} (1+z)^{\frac{1-\mu}{2}}\right).
\end{align*}
The first-order derivative is
\[\frac{\mathrm{d} \bar{\psi}}{\mathrm{d} z} = -\frac{\mu}{1-\mu^2} - \frac{1}{2(1+\mu)} \left(\frac{1-z}{1+z}\right)^{\frac{1+\mu}{2}} + \frac{1}{2(1-\mu)} \left(\frac{1+z}{1-z}\right)^{\frac{1-\mu}{2}},\]
which is zero at $z=0$.
The second-order derivative is
\[\frac{\mathrm{d}^2 \bar{\psi}}{\mathrm{d} z^2} = \frac{1}{1-z^2} \cdot (1-z)^{-\frac{1-\mu}{2}} (1+z)^{-\frac{1+\mu}{2}} \geq 0,\]
which implies two facts: $\bar{\psi}(z)$ tends to zero if and only if $z$ tends to zero, and $\bar{\psi}(z)$ is convex (henceforth $\psi = \bar{\psi}$). Thus, the equivalent loss is classification-calibrated.

From the facts that
\[\frac{\mathrm{d} \psi}{\mathrm{d} z}\Bigg\vert_{z=0} = 0\]
and
\[\frac{\mathrm{d}^2 \psi}{\mathrm{d} z^2}\Bigg\vert_{z=0} = 1,\]
we can say that
\[\psi(z) \sim \frac12 z^2\]
around $z=0$. Due to the boundedness of $\psi$, we have
\[\psi(z) = \Theta(z^2),\]
where the big theta notation is ``of the same order as''.
\end{example}

\subsection{Convexity and Proof of Proposition \ref{prop:Hinge_non-conv}}
In this subsection, we examine how the convexity requirements are fulfilled in the listed examples.
\setcounter{example}{0}
\begin{example}[Convexity of probabilistic model]
W.l.o.g. we still assume $Y=1$ to ease the burden of notations. For the entropy uncertainty, remind that we have already shown its partial derivative with respect to $\hat{Y}$ by
\[\frac{\partial \tilde{\ell}}{\partial \hat{Y}} = \frac12 \log\left(\frac{1+\hat{Y}}{2}\right) + \frac12\cdot \frac{1-\hat{Y}}{1+\hat{Y}} \log\left(\frac{1-\hat{Y}}{2}\right).\]
Continue to compute its partial derivative, we have
\begin{align*}
\frac{\partial^2\tilde\ell}{\partial\hat Y^2}
&=\frac{\partial}{\partial\hat Y}
\left(\frac{\partial\tilde\ell}{\partial\hat Y}\right)\\
&=\frac{\partial U}{\partial\hat Y}\frac{\partial\ell}{\partial\hat Y}
+U\frac{\partial^2\ell}{\partial\hat Y^2}\\
&=\frac{\partial U}{\partial\hat Y}\frac{\partial\ell}{\partial\hat Y},
\end{align*}
which ensures its convexity with respect to $\hat{Y}$.

For the least confidence uncertainty, the partial derivative is
\[\frac{\partial \tilde{\ell}}{\partial \hat{Y}} = \begin{cases}
-\frac12 \cdot \frac{1-\hat{Y}}{1+\hat{Y}}, &\quad \text{if } \hat{Y}\geq 0;\\
-\frac12, & \quad \text{if } \hat{Y}\leq 0,
\end{cases}\]
which implies that $\tilde{\ell}$ is at least $C^1$ continuous with respect to $\hat{Y}$. Furthermore,
\[\frac{\partial^2 \tilde{\ell}}{\partial \hat{Y}^2} = \begin{cases}
\frac{1}{(1+\hat{Y})^2}, &\quad \text{if } \hat{Y}> 0;\\
0, & \quad \text{if } \hat{Y}< 0.
\end{cases}\]
Therefore, the equivalent loss is convex with respect to $\hat{Y}$.

We have shown the convexity with respect to $\hat{Y}$ for both cases. If $\hat{Y}$ is linear with respect to $\theta$, then we can further conclude that the convexity regarding $\theta$ holds. But unlike the margin-based classifiers, the probabilistic models restrict that $\hat{Y} \in (-1, 1)$, where a popular model is the logistic regression model that predicts $\hat{Y} = \frac{\exp(\theta^\top X) - 1}{\exp(\theta^\top X) +1}$. Unlike the original cross-entropy loss, the equivalent loss under the logistic regression model is no longer convex with respect to the parameter $\theta$.

\end{example}

\begin{example}[Convexity of margin-based model]
Since the model is linear in the sense that $\hat{Y} = \theta^\top X$, we only need to check the convexity with respect to $\hat{Y}$. First, assuming $Y=1$, the derivative of $\tilde{\ell}$ with respect to $\hat{Y}$ is
\[\frac{\partial \tilde{\ell}_\mu}{\partial \hat{Y}} = \begin{cases}
\frac{2(\hat{Y} - 1)}{1-\mu \hat{Y}}, &\quad \text{if } \hat{Y} \leq 0;\\
\frac{2(\hat{Y} - 1)}{1+\mu \hat{Y}}, &\quad \text{if } \hat{Y} \in (0, 1);\\
0, &\quad \text{if } \hat{Y} \geq 1.
\end{cases}\]
We can see that $\tilde{\ell}$ is $C^1$ continuous with respect to $\hat{Y}$. We further compute that
\[\frac{\partial^2 \tilde{\ell}_\mu}{\partial \hat{Y}^2} = \begin{cases}
\frac{2(1-\mu)}{(1-\mu \hat{Y})^2}, &\quad \text{if } \hat{Y} < 0;\\
\frac{2(1+\mu)}{(1+\mu \hat{Y})^2}, &\quad \text{if } \hat{Y} \in (0, 1);\\
0, &\quad \text{if } \hat{Y} > 1.
\end{cases}\]
Hence the model is convex but not strongly convex.
\end{example}

\begin{example}[Nonconvexity of threshold-based model]
The equivalent loss is non-convex for $\hat{Y}$ since it is a truncated logistic loss outside a region, where the truncation is to set the loss to be a constant. By the linearity of $\hat{Y}$ on $\theta$, the model is also non-convex for $\theta$.
\end{example}

\begin{example}[Hinge loss with margin uncertainty]
\begin{proof}[Proof of Proposition \ref{prop:Hinge_non-conv}]
It suffices to consider convexity in the scalar prediction. At differentiable points,
\[\frac{\partial^2 \tilde{\ell}}{\partial \hat{Y}^2} = \frac{\partial }{\partial \hat{Y}} \left(\frac{\partial \tilde{\ell}}{\partial \hat{Y}}\right) = \frac{\partial U}{\partial \hat{Y}} \cdot \frac{\partial l}{\partial \hat{Y}} + U \cdot \frac{\partial^2 l}{\partial \hat{Y}^2} = \frac{\partial U}{\partial \hat{Y}} \cdot \frac{\partial l}{\partial \hat{Y}},\]
since the Hinge loss $\ell$ is piecewise linear with respect to $\hat{Y}$. For any fixed $\hat{Y}$, the actual outcome $Y$ could possibly be either $+1$ or $-1$, indicating that
\[\frac{\partial l}{\partial \hat{Y}} = \begin{cases}
+1, &\quad \text{if } \hat{Y} > -1, Y = -1;\\
-1, &\quad \text{if } \hat{Y} < +1, Y = +1;\\
0, &\quad \text{otherwise}.
\end{cases}\]
At the positive part $\hat{Y} > 0$, the uncertainty function is non-increasing, which restricts the term $\frac{\partial U}{\partial \hat{Y}}$ to be non-positive. But for the case $Y=-1$, the convexity requires the term $\frac{\partial U}{\partial \hat{Y}}$ to be non-negative. Henceforth
\[\frac{\partial U}{\partial \hat{Y}} = 0,\]
which implies that the uncertainty function must be piecewise constants. To further ensure that $U$ must be only one constant, we observe that the equivalent loss is now piecewise linear with non-increasing slopes for $\hat{Y} > 0$ if $Y = -1$. In order to keep the loss continuous and convex, the slope must be constant everywhere.
\end{proof}
\end{example}

\begin{example}[Convexity of exponential loss and exponential uncertainty]
Similar to the arguments in Example \ref{eg:raj&bach}, we only need to compute the second-order derivatives (w.l.o.g. assume $Y=1$):
\[\frac{\partial^2\tilde{\ell}_\mu}{\partial \hat{Y}^2} = 
\begin{cases}
(1+\mu) \cdot \exp(-(1+\mu) \hat{Y}), &\quad \text{if } \hat{Y} < 0;\\
(1-\mu) \cdot \exp(-(1-\mu) \hat{Y}), &\quad \text{if } \hat{Y} > 0.
\end{cases}\]
The convexity thus holds.
\end{example}

\subsection{Lipschitzness of Equivalent Loss}
\label{apd:Lipschitz}
Lipschitz condition of those examples in Section \ref{sec:binary} will be verified as follows.

Recall that the Lipschitzness of the equivalent loss is no worse than the original loss by a factor of $M_U$ if $ U \in [0, M_U]$. Moreover, the uncertainty function $U$ is usually decreasing to be near zero when $|\hat{Y}|$ is large enough, which counteracts the effects of the rapid growth of many popular loss functions when $\hat{Y} \cdot Y$ is negative and far enough from zero. To see this, we have a closer look at the probabilistic model in Example \ref{eg:probabilistic}.
\setcounter{example}{0}
\begin{example}[Lipschitzness of probabilistic model]
The original cross-entropy loss is not Lipschitz on the range $\hat{Y} \in (-1, 1)$ (or equivalently, $q \in (0, 1)$), since the derivative of the negative logarithm will explode near the zero point. But from direct computation, for the entropy uncertainty \citep{dagan1995committee}, we have (w.l.o.g. assume $Y=1$)
\allowdisplaybreaks
\begin{align*}
\frac{\partial \tilde{\ell}}{\partial \hat{Y}} &= U \cdot \frac{\partial l}{\partial \hat{Y}}\\
& = -\left[\frac{1+\hat{Y}}{2}\log\left(\frac{1+\hat{Y}}{2}\right)+ \frac{1-\hat{Y}}{2} \log \left(\frac{1-\hat{Y}}{2}\right)\right] \cdot \left(-\frac{1}{1+\hat{Y}}\right)\\
& = \frac12 \log\left(\frac{1+\hat{Y}}{2}\right) + \frac12\cdot \frac{1-\hat{Y}}{1+\hat{Y}} \log\left(\frac{1-\hat{Y}}{2}\right).
\end{align*}
Since the first-order partial derivative $\frac{\partial \tilde{\ell}}{\partial \hat{Y}}$ is non-positive and monotonically increasing, we only need to check the limit case $\hat{Y} \rightarrow -1^+$ to examine the Lipschitzness. We have
\[\bigg|\frac{\partial \tilde{\ell}}{\partial \hat{Y}}\bigg| \sim \frac12 \bigg|\log(1+\hat{Y})\bigg|,\]
which is much smaller than the original loss
\[\bigg|\frac{\partial l}{\partial \hat{Y}}\bigg| \sim \bigg|-\frac{1}{1+\hat{Y}}\bigg|,\]
since by l'H\^{o}pital's rule,
\[\lim_{\hat{Y} \rightarrow -1^+} \frac{\partial \tilde{\ell}}{\partial \hat{Y}}\bigg/ \frac{\partial l}{\partial \hat{Y}} = \lim_{\hat{Y} \rightarrow -1^+} \left(-\frac{1}{1+\hat{Y}}\right)\bigg/\left(-\frac{1}{(1+\hat{Y})^2}\right) = \lim_{\hat{Y} \rightarrow -1^+} (1+\hat{Y}) = 0.\]
Although we cannot say that the equivalent loss is Lipschitz with respect to the whole $(-1, 1)$, for any compact subset of $(-1, 1)$, the equivalent loss is Lipschitz. We shall see that the Lipschitz constant is reduced compared to the original loss.

As for the least confidence uncertainty, the situation is even better: the equivalent loss is Lipschitz over the entire set $\hat{Y} \in (-1, 1)$. To see this, we w.l.o.g. assume $Y = 1$, and the equivalent loss is
\[\tilde{\ell}(\hat{Y} \cdot Y) = \begin{cases}
\frac12 \big(\hat{Y}-2\log(1+\hat{Y})\big), &\quad \text{if } \hat{Y}\geq 0;\\
-\frac12 \cdot \hat{Y}, & \quad \text{if } \hat{Y}\leq 0.
\end{cases}\]
Its partial derivative is
\[\frac{\partial \tilde{\ell}}{\partial \hat{Y}} = \begin{cases}
-\frac12 \cdot \frac{1-\hat{Y}}{1+\hat{Y}}, &\quad \text{if } \hat{Y}\geq 0;\\
-\frac12, & \quad \text{if } \hat{Y}\leq 0,
\end{cases}\]
which implies that the equivalent loss is $\frac12$-Lipschitz.
\end{example}

\begin{example}[Lipschitzness of margin-based model]
From direct computation, the partial derivative with respect to $\hat{Y}$ can be upper-bounded by
\[\left\|\frac{\partial \tilde{\ell}}{\partial \hat{Y}}\right\| \leq \frac{2}{\mu}.\]
By assuming an almost upper bound $M_X$ on the feature space $\mathcal{X}$, the equivalent loss is of course $\frac{2 M_X}{\mu}$-Lipschitz with respect to $\theta$.
\end{example}

\begin{example}[Lipschitzness of threshold-based model]
By the property of the logistic loss, the equivalent loss must be $1$-Lipschitz with respect to $\hat{Y}$. Hence the equivalent loss is $M_X$-Lipschitz with respect to $\theta$.
\end{example}

\begin{example}[Lipschitzness of margin loss and margin-based uncertainty]
The equivalent loss is $1$-Lipschitz with respect to $\hat{Y}$, which indicates its $M_X$-Lipschitzness with respect to $\theta$.
\end{example}

\begin{example}[Lipschitzness of exponential loss and exponential uncertainty]
The prediction $\hat{Y} = \theta^\top X$ has an upper bound of
\[|\hat{Y}| \leq M_X \cdot M_{\Theta},\]
where $M_X$ is the almost sure upper bound for $X$ and $M_{\Theta}$ is the upper bound for $\Theta$. Then the equivalent loss has an upper bound for its partial derivative with respect to $\hat{Y}$ of $\exp\big((1-\mu) M_X \cdot M_{\Theta}\big)$. The final Lipschitzness constant with respect to $\theta$ is $M_X \cdot \exp\big((1-\mu) M_X \cdot M_{\Theta}\big)$.
\end{example}

\section{Proofs and Discussions}

\subsection{Proof of Proposition \ref{prop:SGD_equiv}}
\begin{proof}
Use the two-stage filtration defined in Section~\ref{sec:binary}. Conditional on $\mathcal F_t^X\vee\sigma(Y_t)$, both $U_t$ and the sample gradient are fixed. Independence of $\xi_t$ gives
\[
\E[\mathbbm{1}\{\xi_t\le U_t\}\mid\mathcal F_t^X\vee\sigma(Y_t)]=U_t.
\]
Consequently, the update and the equivalent-loss relation imply
\begin{align*}
\E[\theta_{t+1}-\theta_t\mid\mathcal F_t^X\vee\sigma(Y_t)]
&=-\eta_t U_t\nabla_\theta\ell(\theta_t;(X_t,Y_t))\\
&=-\eta_t\nabla_\theta\tilde\ell(\theta_t;(X_t,Y_t)).
\end{align*}
A further expectation over the current feature and label gives the population-gradient identity for an $\mathcal F_t$-measurable learning rate, including the fixed rate used in Proposition~\ref{prop:SGD_converge}.
\end{proof}

\subsection{Proof of Proposition \ref{prop:SGD_converge}}
\label{apd:constant-c-proof}
\begin{proof}
Write $R(\theta)=\E_{(X,Y)\sim\mathcal P}[\tilde\ell(\theta;(X,Y))]$ as in the original population-risk notation, and let
\[
g_t=\mathbbm{1}\{\xi_t\le U_t\}\nabla_\theta\ell(\theta_t;(X_t,Y_t)).
\]
With the two-stage filtration from Section~\ref{sec:binary}, independence of the query randomization gives
\begin{align*}
\E[g_t\mid\mathcal F_t^X\vee\sigma(Y_t)]
&=U_t\nabla_\theta\ell(\theta_t;(X_t,Y_t)),\\
\E[g_t\mid\mathcal F_t]
&=\nabla R(\theta_t).
\end{align*}
Because $\eta_t=cD/(G\sqrt T)$ is deterministic,
\[
\E[\eta_t(g_t-\nabla R(\theta_t))\mid\mathcal F_t]=0.
\]
For the second moment, conditioning first on the feature and using \eqref{eq:feature-second-moment} yields
\begin{align*}
\E[\|g_t\|_2^2\mid\mathcal F_t^X]
&=U_t\E[\|\nabla_\theta\ell(\theta_t;(X_t,Y_t))\|_2^2
\mid\mathcal F_t^X]\le G^2U_t,\\
\E[\|g_t\|_2^2\mid\mathcal F_t]
&\le G^2\E[U_t\mid\mathcal F_t].
\end{align*}
In particular, the factor $\E[U_t\mid\mathcal F_t]$ is retained without treating $U_t$ and the unqueried gradient norm as independent.

Expanding the update, taking conditional expectation, and using convexity of $R$ gives
\begin{align*}
\E[\|\theta_{t+1}-\tilde\theta^*\|_2^2\mid\mathcal F_t]
&\le\|\theta_t-\tilde\theta^*\|_2^2
-2\eta_t\{R(\theta_t)-R(\tilde\theta^*)\}\\
&\quad+\eta_t^2G^2\E[U_t\mid\mathcal F_t].
\end{align*}
Set $\eta=cD/(G\sqrt T)$. Summing, taking total expectations, and discarding the final nonnegative squared distance gives
\[
2\eta\sum_{t=1}^T\E[R(\theta_t)-R(\tilde\theta^*)]
\le D^2+\eta^2G^2\E[Q_T].
\]
The weights in Algorithm~\ref{alg:USGD} are now all equal. Convexity at $\bar\theta_T=T^{-1}\sum_t\theta_t$ therefore implies
\begin{align*}
\E[R(\bar\theta_T)-R(\tilde\theta^*)]
&\le \frac{D^2}{2\eta T}
+\frac{\eta G^2}{2T}\E[Q_T]\\
&=\frac{DG}{2\sqrt T}
\left(\frac1c+c\E[r_T]\right).
\end{align*}
Every denominator in this argument is deterministic. Finally,
\[
\E\!\left[\sum_{t=1}^T\mathbbm{1}\{\xi_t\le U_t\}\right]
=\sum_{t=1}^T\E[U_t]=\E[Q_T],
\]
which establishes the interpretation of the expected label budget.
\end{proof}

\paragraph{A weaker moment assumption.}
If only $\E[\|\nabla\ell(\theta_t;(X_t,Y_t))\|_2^2\mid\mathcal F_t]\le G^2$ is assumed, the same proof yields
\[
\E[R(\bar\theta_T)-R(\tilde\theta^*)]
\le\frac{DG}{2\sqrt T}\left(\frac1c+cU_{\text{max}}\right),
\]
because $U_t\le U_{\text{max}}$ can be used pointwise. With $c=U_{\text{max}}^{-1/2}$, this is $DG\sqrt{U_{\text{max}}}/\sqrt T$. Under the same link and approximation assumptions as Corollary~\ref{coro:final_sample_complexity}, the matched leading binary-risk ratio is $(\E[r_T]/U_{\text{max}})^{1/4}\le1$. The feature-conditional moment bound in Proposition~\ref{prop:SGD_converge} retains the average rate inside the variance term and improves this coefficient by the factor $\sqrt{(1+\E[r_T]/U_{\text{max}})/2}$.

\subsection{Proof of Theorem \ref{thm:stream-based}}
\begin{proof}
Apply the surrogate-risk inequality to each realized $f_{\bar\theta_T}$ and take expectations. Jensen's inequality for the convex link $\psi$ yields
\begin{align*}
&\psi\!\left(\E[L_{\mathrm{01}}(f_{\bar\theta_T})]
-\inf_{g\in\mathcal G}L_{\mathrm{01}}(g)\right)\\
&\quad\le\E[\tilde\ell(f_{\bar\theta_T}(X),Y)]
-\inf_{g\in\mathcal G}\E[\tilde\ell(g(X),Y)]\\
&\quad=\E[R(\bar\theta_T)-R(\tilde\theta^*)]
+\inf_{\theta\in\Theta}\E[\tilde\ell(f_\theta(X),Y)]
-\inf_{g\in\mathcal G}\E[\tilde\ell(g(X),Y)].
\end{align*}
Use Proposition~\ref{prop:SGD_converge} to bound the first term and apply the nondecreasing generalized inverse of $\psi$.
\end{proof}

\subsection{Proof of Corollary \ref{coro:final_sample_complexity}}
\label{apd:comparison-proof}
\begin{proof}
Proposition~\ref{prop:equiv_surrogate} and $-H''(1/2)>0$ give
\[
\tilde\psi^{-1}(u)
=\left(\frac{8u}{U_{\text{max}}[-H''(1/2)]}\right)^{1/2}(1+o(1)),
\qquad u\downarrow0.
\]
Substitute \eqref{eq:constant-c-bound} in Theorem~\ref{thm:stream-based}, with zero approximation error. This proves \eqref{eq:fixed-c-classification}. For passive SGD using $n$ labels, the ordinary fixed-step bound is $DG/\sqrt n$, and the original-loss link gives leading binary-risk bound
\[
\left(\frac{8DG}{[-H''(1/2)]\sqrt n}\right)^{1/2}.
\]
Here the same $D$ must bound distance to the respective population minimizers, and the same $G$ must satisfy the active and passive gradient bounds. A bound on the diameter of $\Theta$, together with uniformly bounded sample gradients, is one sufficient common choice.

Set $n=\E[Q_T]=T\E[r_T]$. If $n$ is not an integer, take the adjacent integer label budget; the resulting factor tends to one as $n\to\infty$. Dividing the two leading bounds gives
\[
\left[
\frac{\sqrt{\E[r_T]}}{2U_{\text{max}}}
\left(\frac1c+c\E[r_T]\right)
\right]^{1/2}.
\]
Substitution of $c=U_{\text{max}}^{-1/2}$ proves \eqref{eq:feasible-risk-ratio}. Since $0<\E[r_T]/U_{\text{max}}\le1$, both factors in that expression are at most one and are strictly smaller than one when the ratio is below one. Under the oracle relation $c^{-2}=\E[r_T]$, the same calculation gives $(\E[r_T]/U_{\text{max}})^{1/2}$.
\end{proof}

\subsection{Fixed-step calibration and label-budget interpretation}
\label{apd:step-calibration}
The constant in \eqref{eq:constant-c-bound} makes the role of rate information explicit. Its worst-case value over $0\le\E[r_T]\le U_{\text{max}}$ is $c^{-1}+cU_{\text{max}}$, minimized at $c=U_{\text{max}}^{-1/2}$. At matched expected label usage, the squared leading-risk ratio \eqref{eq:fixed-c-risk-ratio} is also increasing in $\E[r_T]$ for each fixed $c$. Its supremum is
\[
\frac12\left(c\sqrt{U_{\text{max}}}
+\frac{1}{c\sqrt{U_{\text{max}}}}\right),
\]
whose minimum is one at the same choice. Thus this choice is optimal for the worst-case comparison of these upper bounds when only the range of the query rate is known.

If a positive rate estimate $\widehat r$ is fixed before the analyzed run and $c=\widehat r^{-1/2}$, the equivalent-risk bound is
\[
\frac{DG}{2\sqrt T}
\left(\sqrt{\widehat r}+\frac{\E[r_T]}{\sqrt{\widehat r}}\right).
\]
Relative to the balanced oracle expression for the same rate, the multiplicative inflation is
\[
\frac12\left(
\sqrt{\frac{\widehat r}{\E[r_T]}}
+\sqrt{\frac{\E[r_T]}{\widehat r}}\right).
\]
For example, a multiplicative estimate within a factor $K\ge1$ yields inflation at most $(\sqrt K+1/\sqrt K)/2$. This is a calibration sensitivity statement: changing $c$ can itself change the expected query rate, so the rate in every formula belongs to the run being analyzed. A realized terminal $r_T$ cannot be substituted retrospectively for a deterministic tuning constant. A pilot estimate may instead be conditioned on at the start of the subsequent run. Sliding-window estimates motivate such calibration, while continuously changing the step along the same run requires a separate analysis of the chosen averaging rule.

Risk ratios and budget ratios describe different comparisons. At a fixed expected label budget, the oracle and feasible leading classification-risk ratios are, respectively,
\[
\left(\frac{\E[r_T]}{U_{\text{max}}}\right)^{1/2},\qquad
\left(\frac{\E[r_T]}{U_{\text{max}}}\right)^{1/4}
\sqrt{\frac{1+\E[r_T]/U_{\text{max}}}{2}}.
\]
For a regime in which the rate ratio approaches a fixed positive value $\rho$ and all other leading constants are fixed, the bounds scale as $n^{-1/4}$ in the expected label budget $n$. At the same target accuracy, their leading sufficient-budget ratios are consequently the fourth powers:
\[
\rho^2\quad\text{(oracle)},\qquad
\frac{\rho(1+\rho)^2}{4}\quad\text{(feasible)}.
\]
These are implications of the leading bounds. When the query rate changes with the horizon, its horizon dependence must also be retained when solving for a label budget.

\subsection{Proof of Proposition \ref{prop:equiv_surrogate}}
\begin{proof}
Let $\tilde C_p(s)=p\tilde\ell(s)+(1-p)\tilde\ell(-s)$ and $\tilde H(p)=\inf_s\tilde C_p(s)$. Evenness and positivity of $U$ give
\[
\tilde C_p'(s)=U(s)\{p\ell'(s)-(1-p)\ell'(-s)\}=U(s)C_p'(s).
\]
Because $C_p$ is convex with a unique minimizer $s^*(p)$, the two derivatives have the same sign on either side of that minimizer. Thus $s^*(p)$ also uniquely minimizes $\tilde C_p$.

The envelope theorem yields
\[
H'(p)=\ell(s^*(p))-\ell(-s^*(p)),\qquad
\tilde H'(p)=\tilde\ell(s^*(p))-\tilde\ell(-s^*(p)).
\]
Differentiating and using the defining weighted derivative and evenness of $U$,
\begin{align*}
\tilde H''(p)
&=\{\tilde\ell'(s^*(p))+\tilde\ell'(-s^*(p))\}\,{s^*}'(p)\\
&=U(s^*(p))\{\ell'(s^*(p))+\ell'(-s^*(p))\}\,{s^*}'(p)\\
&=U(s^*(p))H''(p).
\end{align*}
Symmetry and uniqueness imply $s^*(1/2)=0$, hence $H'(1/2)=\tilde H'(1/2)=0$ and $\tilde H''(1/2)=U_{\text{max}}H''(1/2)$.

For $p\ge1/2$, $C_p'(0)=(2p-1)\ell'(0)\le0$. Convexity places the minimum on the wrong-sign half-line at zero. The sign identity for $\tilde C_p'$ gives the same boundary minimizer for the equivalent loss; the case $p\le1/2$ is symmetric. Consequently $H^-(p)=\ell(0)$ and $\tilde H^-(p)=\tilde\ell(0)$. Both $H$ and $\tilde H$ are concave as infima of affine functions of $p$. Their link candidates, a constant minus the corresponding Bayes risk at $(1+z)/2$, are therefore convex and equal their biconjugates. Expanding at $p=1/2$ gives
\[
\psi(z)=-\frac18H''(1/2)z^2+o(z^2),\qquad
\tilde\psi(z)=-\frac{U_{\text{max}}}{8}H''(1/2)z^2+o(z^2).
\]
\end{proof}

\subsection{Proof of Proposition \ref{prop:existence_equiv}}
\label{subapd:proof_existence_equiv}

\begin{proof}
Let
\[
F(\theta):=U(\theta) \cdot \nabla_\theta\ell(\theta),
\qquad
F_i(\theta):=U(\theta) \cdot \partial_i\ell(\theta).
\]

\noindent\textbf{(1) $\Rightarrow$ (2).}
If $\nabla_\theta\tilde{\ell}=F$, then $F_i=\partial_i\tilde{\ell}$ for all $i$, so by equality of
mixed partial derivatives,
\[
\partial_jF_i=\partial_iF_j
\qquad
\text{for all } i,j\in[k].
\]
Expanding both sides yields
\[
\partial_jU\,\partial_i\ell+U\,\partial_{ji}\ell
=
\partial_iU\,\partial_j\ell+U\,\partial_{ij}\ell.
\]
Since $\ell\in C^2$, we have $\partial_{ij}\ell=\partial_{ji}\ell$, hence
\[
\partial_jU\,\partial_i\ell
=
\partial_iU\,\partial_j\ell.
\]
This is exactly condition (2).

\medskip
\noindent\textbf{(2) $\Rightarrow$ (1).}
Using $F_i=U\,\partial_i\ell$, we compute
\[
\partial_jF_i-\partial_iF_j
=
(\partial_jU)(\partial_i\ell)-(\partial_iU)(\partial_j\ell)
+
U(\partial_{ji}\ell-\partial_{ij}\ell).
\]
By condition (2) and the symmetry of mixed partials, we obtain
\[
\partial_jF_i=\partial_iF_j
\qquad
\text{for all } i,j\in[k].
\]
Thus $F$ is curl-free. Since $\Theta$ is open and simply connected, the standard path-independence
criterion for conservative vector fields implies that there exists $\tilde{\ell}\in C^2(\Theta)$ such that
\[
\nabla_\theta\tilde{\ell}=F=U\,\nabla_\theta\ell.
\]

\medskip
\noindent\textbf{(2) $\Rightarrow$ (3).}
Fix $\theta\in\Theta$ with $\nabla_\theta\ell(\theta)\neq 0$. Choose $m\in[k]$ such that
$\partial_m\ell(\theta)\neq 0$. By continuity, after shrinking to a neighborhood of $\theta$ if
necessary, we may assume $\partial_m\ell\neq 0$ on that neighborhood. Define
\[
\lambda:=\frac{\partial_mU}{\partial_m\ell}.
\]
Then for any $i\in[k]$, condition (2) with the pair $(i,m)$ gives
\[
\partial_iU\,\partial_m\ell=\partial_mU\,\partial_i\ell,
\]
hence
\[
\partial_iU=\lambda\,\partial_i\ell.
\]
Therefore
\[
\nabla_\theta U=\lambda\,\nabla_\theta\ell.
\]
This proves (3).

\medskip
\noindent\textbf{(3) $\Rightarrow$ (2).}
Fix $\theta\in\Theta$.

If $\nabla_\theta\ell(\theta)=0$, then
\[
\partial_iU(\theta)\,\partial_j\ell(\theta)
=
\partial_jU(\theta)\,\partial_i\ell(\theta)
=
0
\qquad
\text{for all } i,j\in[k],
\]
so (2) holds trivially at $\theta$.

If $\nabla_\theta\ell(\theta)\neq 0$, then by (3) there exists $\lambda(\theta)$ such that
\[
\partial_iU(\theta)=\lambda(\theta)\,\partial_i\ell(\theta)
\qquad
\text{for all } i\in[k].
\]
Hence
\[
\partial_iU(\theta)\,\partial_j\ell(\theta)
=
\lambda(\theta)\,\partial_i\ell(\theta)\,\partial_j\ell(\theta)
=
\partial_jU(\theta)\,\partial_i\ell(\theta).
\]
Thus (2) holds at every $\theta\in\Theta$.

\medskip
\noindent\textbf{(3) $\Rightarrow$ (4).}
Fix $c\in\mathbb{R}$ and let $C$ be a path-connected component of $L_c^{\mathrm{reg}}$.
We show that $U$ is locally constant on $C$.

Take any $\theta^0\in C$. Since $\nabla_\theta\ell(\theta^0)\neq 0$, after relabeling coordinates we
may assume $\partial_k\ell(\theta^0)\neq 0$. By the implicit function theorem, there exist an open
connected set $B\subset\mathbb{R}^{k-1}$, an open neighborhood $W$ of $\theta^0$, and a $C^1$
function $g:B\to\mathbb{R}$ such that
\[
L_c^{\mathrm{reg}}\cap W
=
\{(x,g(x)):x\in B\},
\]
where $x=(\theta_1,\dots,\theta_{k-1})$.

Define
\[
h(x):=U(x,g(x)).
\]
Since $\ell(x,g(x))\equiv c$ on $B$, differentiating with respect to $x_i$ gives
\[
\partial_i\ell(x,g(x))
+
\partial_k\ell(x,g(x))\,\partial_i g(x)
=
0,
\]
and therefore
\[
\partial_i g(x)
=
-\frac{\partial_i\ell}{\partial_k\ell}(x,g(x)),
\qquad i=1,\dots,k-1.
\]
By condition (3), on $L_c^{\mathrm{reg}}\cap W$ we have
\[
\partial_iU=\lambda\,\partial_i\ell,
\qquad
\partial_kU=\lambda\,\partial_k\ell
\]
for some scalar field $\lambda$. Hence
\[
\partial_i h
=
\partial_iU+\partial_kU\,\partial_i g
=
\lambda\,\partial_i\ell
-
\lambda\,\partial_k\ell\cdot\frac{\partial_i\ell}{\partial_k\ell}
=
0.
\]
Thus $h$ is constant on $B$, so $U$ is constant on $L_c^{\mathrm{reg}}\cap W$.
Therefore $U|_C$ is locally constant. Since $C$ is path-connected, hence connected, a locally
constant function on $C$ must be constant. This proves (4).

\medskip
\noindent\textbf{(4) $\Rightarrow$ (3).}
Fix $\theta^0\in\Theta$ such that $\nabla_\theta\ell(\theta^0)\neq 0$, and set
$c:=\ell(\theta^0)$. After relabeling coordinates, assume $\partial_k\ell(\theta^0)\neq 0$.
By the implicit function theorem, there exist an open connected set $B\subset\mathbb{R}^{k-1}$,
an open neighborhood $W$ of $\theta^0$, and a $C^1$ function $g:B\to\mathbb{R}$ such that
\[
L_c^{\mathrm{reg}}\cap W
=
\{(x,g(x)):x\in B\}.
\]
Since $B$ is open and connected in $\mathbb{R}^{k-1}$, it is path-connected; hence its graph
is path-connected as well. By condition (4), the function
\[
h(x):=U(x,g(x))
\]
is constant on $B$. Therefore, for each $i=1,\dots,k-1$,
\[
0=\partial_i h=\partial_iU+\partial_kU\,\partial_i g.
\]
Using
\[
\partial_i g=-\frac{\partial_i\ell}{\partial_k\ell},
\]
we obtain
\[
\partial_iU
=
\frac{\partial_kU}{\partial_k\ell}\,\partial_i\ell,
\qquad i=1,\dots,k-1.
\]
If we define
\[
\lambda:=\frac{\partial_kU}{\partial_k\ell},
\]
then the same identity also holds trivially for $i=k$. Therefore
\[
\nabla_\theta U=\lambda\,\nabla_\theta\ell
\]
at $\theta^0$. Since $\theta^0$ was arbitrary among points with $\nabla_\theta\ell\neq 0$, condition
(3) follows.
\end{proof}

\subsection{Proof of Proposition \ref{prop:SGD_converge_pool}}
\begin{proof}
Condition on the pool and let $\hat R(\theta)=\hat{\tilde\ell}(\theta)$ and $g_t=\nabla_\theta\ell(\theta_t;(X_{i_t},Y_{i_t}))$. The pre-index history includes the pool, $\theta_t$, and $S_t$, so
\[
\E[g_t\mid\mathcal F_t]=S_t^{-1}\nabla\hat R(\theta_t).
\]
For deterministic $\eta=D/(G\sqrt T)$, the squared-distance expansion and conditional gradient bound give
\[
\E\!\left[\sum_{t=1}^T\frac{\eta}{S_t}
\{\hat R(\theta_t)-\hat R(\hat{\tilde\theta}^*)\}\right]
\le\frac{D^2}{2}+\frac{T\eta^2G^2}{2}.
\]
Every excess-risk summand is nonnegative because $\hat{\tilde\theta}^*$ minimizes the fixed pool objective. The weight sum satisfies $\sum_t\eta/S_t\ge\eta T/U_{\text{max}}$ pathwise. Convexity at the original weighted output therefore implies
\begin{align*}
\hat R(\bar\theta_T)-\hat R(\hat{\tilde\theta}^*)
&\le \frac{\sum_t(\eta/S_t)
\{\hat R(\theta_t)-\hat R(\hat{\tilde\theta}^*)\}}
{\sum_t\eta/S_t}\\
&\le\frac{U_{\text{max}}}{\eta T}
\sum_t\frac{\eta}{S_t}
\{\hat R(\theta_t)-\hat R(\hat{\tilde\theta}^*)\}.
\end{align*}
Take expectations and substitute the preceding bound to obtain
\[
\E[\hat R(\bar\theta_T)-\hat R(\hat{\tilde\theta}^*)]
\le U_{\text{max}}\left(\frac{D^2}{2\eta T}+\frac{\eta G^2}{2}\right)
=\frac{DGU_{\text{max}}}{\sqrt T}.
\]
This argument bounds the nonnegative weighted risk sum before expectation; no martingale term is divided by a future-dependent random denominator. In the repeated-query model, replace the labeled empirical objective by its feature-conditional counterpart and use the same argument.
\end{proof}

\section{Details of the Query-Clock Momentum Limit}
\label{apd:momentum-details}
\begin{proof}[Proof of Proposition~\ref{prop:embedded-query-chain}]
Fix the entire optimizer state and its common uncertainty/gradient evaluation point $\theta$. For a measurable set $B$, independence of future raw observations gives
\begin{align*}
\Pr((X,Y)_{\mathrm{next\ query}}\in B\mid\theta)
&=\sum_{j=1}^{\infty}(1-r(\theta))^{j-1}
\int_B U(\theta;x)\,\mathcal P(\mathrm dx,\mathrm dy)\\
&=\frac1{r(\theta)}\int_B U(\theta;x)\,\mathcal P(\mathrm dx,\mathrm dy).
\end{align*}
Applying the equivalent-loss identity inside this integral gives the mean-gradient formula.
\end{proof}

\begin{proof}[Proof of Proposition~\ref{prop:momentum-ode}]
One explicit scaling uses queried observations $(X_{k+1}^Q,Y_{k+1}^Q)$ with conditional law $\mathcal P^Q_{\theta_k}$ and recursion
\begin{align*}
v_{k+1}&=v_k-h a(s_k)v_k-h\nabla\ell(\theta_k;(X_{k+1}^Q,Y_{k+1}^Q)),\\
\theta_{k+1}&=\theta_k+h v_{k+1},\qquad s_k=s_0+kh.
\end{align*}
Take convergent initial conditions, $s_0>0$ when $a(s)=3/s$, and $a$ continuous on the interval of interest. The mean gradient is the normalized equivalent-risk gradient by the embedded-chain result. Subtract this conditional mean from each gradient to obtain a martingale-difference sequence. On a compact set where the conditional second moment is bounded, the quadratic variation of its cumulative $h$-scaled sum over a fixed time interval is $O(h)$. Doob's inequality therefore makes this noise vanish uniformly in probability.

The remaining recursion is an Euler approximation of the first-order position--velocity system corresponding to \eqref{eq:momentum-ode}. Its deterministic discretization error vanishes as $h\to0$. Local Lipschitz continuity, localization to the compact set, and Gronwall's inequality imply uniform convergence in probability of the interpolated trajectory. For Nesterov's recurrence, evaluating both uncertainty and gradient at its common look-ahead point changes the drift evaluation by $O(h)$ on a compact set, yielding the same limiting force and damping $3/s$. A constant solution has zero velocity and zero normalized force; positivity of $r$ makes its stationary condition exactly the one stated.
\end{proof}

The energy identity follows by differentiating the displayed $\mathcal E$ in the main text:
\begin{align*}
\dot{\mathcal E}
&=\nabla\E[\tilde\ell(\theta;(X,Y))]^\top\dot\theta
+\frac12\dot r\|\dot\theta\|^2+r\dot\theta^\top\ddot\theta\\
&=-\left(ar-\frac12\dot r\right)\|\dot\theta\|^2.
\end{align*}
Under the time change $x(t)=\theta(s(t))$ with $\dot s=\sqrt{r(x(t))}$, the ODE equivalently becomes
\[
\ddot x+
\left(a(s(t))\sqrt{r(x)}-\frac12\frac{\mathrm d}{\mathrm dt}\log r(x)\right)\dot x
+\nabla\E[\tilde\ell(x;(X,Y))]=0.
\]
The changing query clock thus enters the damping coefficient when the positional force is written as the gradient of the original equivalent risk.

At a critical point $\theta^*$, the derivative of $r^{-1}$ multiplies the vanishing gradient. Hence the force Jacobian equals $r(\theta^*)^{-1}$ times the population Hessian. With constant damping $\gamma>0$, an eigenvalue $\lambda_i$ of that Hessian gives a scalar mode
\[
\ddot z_i+\gamma\dot z_i+\frac{\lambda_i}{r(\theta^*)}z_i=0.
\]
Positive curvature gives characteristic roots with negative real parts, with critical damping $2\sqrt{\lambda_i/r(\theta^*)}$. A negative eigenvalue gives a positive real root and local instability. Degenerate directions need higher-order analysis.

\section{Additional Related Work}
\label{apd:additional-related-work}

Several adjacent literatures help place the equivalent-loss perspective within active learning. Statistical approaches to acquisition select observations by their anticipated effect on predictive error or variance \citep{cohn1996statistical}. Disagreement-based agnostic learning instead identifies regions in which plausible hypotheses disagree \citep{balcan2006agnostic,hanneke2014theory}. Importance-weighted active learning couples adaptive querying with a correction for sampling bias and controls the resulting variance to obtain label-complexity guarantees \citep{beygelzimer2009importance}. This provides a useful conceptual comparison: importance weighting preserves the original risk, while the equivalent-loss construction identifies the objective induced when the query weight remains in the expected gradient.

Recent theory has refined the attainable benefits of active learning beyond these earlier characterizations. \citet{hanneke2024universal} characterize distribution-dependent learning rates in the universal-learning setting, where guarantees are evaluated for each fixed data distribution. In the agnostic setting, \citet{hanneke2025agnostic} establishes a sharp first-order query-complexity bound whose leading term removes the disagreement-dependent factors of earlier general bounds. These results characterize what suitable acquisition strategies can achieve; the complementary algorithmic question studied here is how a specified uncertainty score and gradient update jointly determine the learning objective and its statistical properties.

A second connection concerns the information represented by an uncertainty score. Deep Bayesian active learning uses posterior predictive uncertainty for data acquisition \citep{gal2017deep}, while BADGE combines uncertainty and batch diversity through gradient embeddings \citep{ash2020deep}. More recently, \citet{rahmati2024mismatch} examine how model mismatch can make uncertainty-based acquisition less effective than random sampling, and \citet{bui2025calibrated} use estimates of predictive calibration error to guide pool-based acquisition. These studies emphasize the interaction between model capacity, confidence, and informative queries. Predictive calibration concerns agreement between reported probabilities and observed frequencies; the surrogate calibration used in our analysis connects population risk minimization to the Bayes decision \citep{bartlett2006convexity}. Both perspectives help assess uncertainty-based learning, at different points between acquisition and prediction.

\end{document}